\documentclass[10pt,journal,compsoc]{IEEEtran}
\usepackage[normalem]{ulem}
\usepackage[usenames, dvipsnames]{color}
\usepackage{xcolor}
\usepackage{booktabs}

\newcommand{\viewname}[1]{\textsc{#1}}

\definecolor{mypink}{HTML}{FFE7E6}
\definecolor{myyellow}{HTML}{FFF8D9}
\definecolor{mygreen}{HTML}{D9FFE7}
\definecolor{mypurple}{HTML}{E6E6FF}

\newcommand{\task}[1]{\colorbox{mypink}{#1}}
\newcommand{\op}[1]{#1}
\newcommand{\ob}[1]{#1}
\newcommand{\infer}[1]{#1}

\newcommand{\nt}[1]{{#1}}

\ifCLASSOPTIONcompsoc
  \usepackage[nocompress]{cite}
\else
  \usepackage{cite}
\fi
\ifCLASSINFOpdf
   \usepackage[pdftex]{graphicx}
\else
   \usepackage[dvips]{graphicx}
\fi
\usepackage{url}
\usepackage{hyperref}

\begin{document}
%
\title{Visualizing Graph Neural Networks with CorGIE: \underline{Cor}responding a \underline{G}raph to \underline{I}ts \underline{E}mbedding}
%
%
%
%

\author{Zipeng~Liu,
  Yang~Wang,
  J{\"u}rgen~Bernard,
  Tamara~Munzner 
  \IEEEcompsocitemizethanks{
    \IEEEcompsocthanksitem Z. Liu and T. Munzner are with the Department of Computer Science, University of British Columbia. Email: {zipeng, tmm}@cs.ubc.ca \protect\\
    \IEEEcompsocthanksitem Yang Wang is with Facebook Inc.  Email: yvv@fb.com \protect\\
    \IEEEcompsocthanksitem J. Bernard is with the Department of Computer Science, University of Zurich. Email: bernard@ifi.uzh.ch}
  \thanks{Manuscript received June 15, 2021; revised XXX.}
}

%
%

\markboth{IEEE TRANSACTIONS ON VISUALIZATION AND COMPUTER GRAPHICS, ~Vol.~xx, No.~x, xxxx~xxxx}%
{Author \MakeLowercase{\textit{et al.}}: Visualizing Graph Neural Networks with CorGIE: Corresponding a Graph to Its Embedding}
%



\IEEEtitleabstractindextext{%
  \begin{abstract}
    Graph neural networks (GNNs) are a class of powerful machine learning tools that model node relations for making predictions of nodes or links.
    GNN developers rely on quantitative metrics of the predictions to evaluate a GNN, but similar to many other neural networks, it is difficult for them to understand if the GNN truly learns characteristics of a graph as expected.
    We propose an approach to corresponding an input graph to its node embedding (aka latent space), a common component of GNNs that is later used for prediction.
    We abstract the data and tasks, and develop an interactive multi-view interface called CorGIE to instantiate the abstraction.  As the key function in CorGIE, we propose the K-hop graph layout to show topological neighbors in hops and their clustering structure.
    To evaluate the functionality and usability of CorGIE, we present how to use CorGIE in two usage scenarios, and conduct a case study with \nt{five} GNN experts.
    
    \textbf{Availability:} Open-source code at \url{https://github.com/zipengliu/CorGIE/}, supplemental materials \& video at \url{https://osf.io/tr3sb/}.
  \end{abstract}

  }

\maketitle

\IEEEdisplaynontitleabstractindextext

%

\IEEEraisesectionheading{\section{Introduction}\label{sec:introduction}}

\IEEEPARstart{G}{raph} neural networks (GNNs) are machine learning (ML) models that have received substantial recent attention due to their ability to deal with abstract concepts like relationships and interactions. GNN models are widely used in downstream ML applications such as fraud detection, traffic modeling, and product recommendation, in addition to the classic ML application domains of natural language processing and computer vision.

\begin{figure*}[!t]
  \centering
  \includegraphics[width=\linewidth]{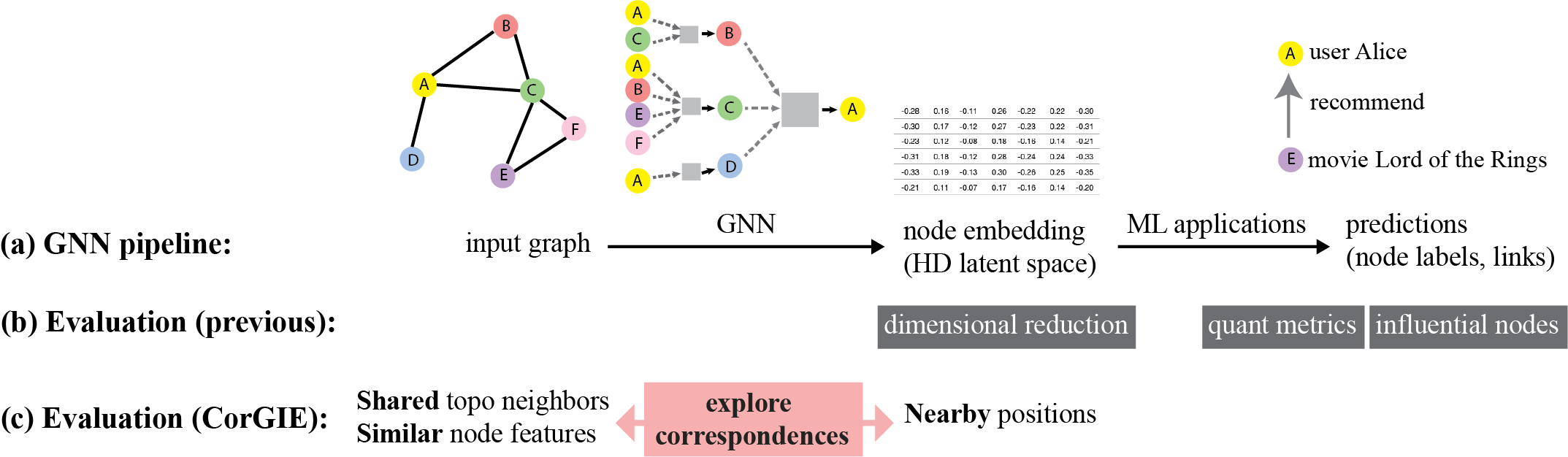}
  \caption[Motivation of CorGIE]{
    CorGIE motivation. (a) Standard pipeline for graph neural network (GNN) training and usage\nt{, with visual examples on top}. (b) Previous approaches to evaluate GNNs mostly focus on the predictions~\cite{tang2015line, ying2019gnnexplainer, dwivedi2020benchmarking}, or to separately inspect the node embedding; (c) our approach with CorGIE is to directly support exploration of correspondences between an input graph and its node embedding.}
  \label{fig:gnn}
\end{figure*}

During training, a GNN model takes in a node-link graph and as output generates a \textbf{node embedding}; 
that is, a representation of all  discrete nodes in the graph as fixed-dimensional vectors in a continuous \textbf{latent space}. Proximities between embedded nodes in the latent space represent meaningful similarities between nodes in the input graph learned during the training. The node embedding leverages both the topology -- connectivity between neighboring nodes -- and node features -- information associated with each node -- 
of the input graph. (Information associated with nodes, typically called node features in the ML literature, is sometimes called node attributes in the visualization literature.) 
The node embedding is then available for feeding into downstream ML applications, for example to make predictions about nodes and links. 
Fig.~\ref{fig:gnn}a summarizes the standard GNN pipeline.

To achieve optimal performance, model developers must evaluate whether the training has succeeded in producing a GNN model that has in fact learned the important characteristics from the input graph. Model developers often need to determine the stopping criteria for training a GNN model; in other words, to answer the question ``are we there yet?'' when training models or tuning hyperparameters.
In addition to global questions such as when to stop training and tuning, 
model developers need to debug and trace errors and sub-optimal states in the trained model.  For example, they may want to identify and understand the groups of nodes that suffer from mis-classification the most.
In some cases, they also need to compare results from different model architectures or hyperparameters to choose the best ones for production use.

The most prevalent current approach for model evaluation is to compute quantitative metrics based on the downstream ML applications~\cite{dwivedi2020benchmarking},
such as precision and F1 for node classification and Hit@k for link prediction, which are used for cross-validation during training. 
Some researchers have proposed algorithms to compute explanations for prediction results in terms of influential nodes, or what features are important 
in the determination of a specific node~\cite{ying2019gnnexplainer}.
Another common approach is to conduct qualitative sanity checks, such as manual inspection of node or link instances, for example by visualizing with multi-class scatterplots of the dimensionally reduced node embedding~\cite{tang2015line}. Fig.~\ref{fig:gnn}b summarizes these previous approaches to evaluation. 

However, these existing evaluation approaches fall short because they do not sufficiently support developers in directly understanding the correspondences between input graphs and their node embeddings. Developers must contend with a GNN model as a black box, because it is hard to check whether all important information is codified in the embedding.  Even if they find errors in the downstream predicted node labels, it is hard to even trace these errors back to the upstream GNNs, or to refine the GNNs to avoid them.

Our key idea is to aid GNN evaluation by surfacing the correspondences, or lack thereof, between an input graph and its node embedding produced by a GNN. Fig.~\ref{fig:gnn}c illustrates this approach. 
Geometrically speaking, a GNN should learn to place graph nodes with similar neighbors and feature values in a nearby location in the high dimensional latent space.
\nt{Therefore, we can evaluate a GNN by verifying how well it preserves the characteristics of the input graph in the node embedding.}
For instance, one can explore how a cluster in the node embedding corresponds to the similarity of topology and/or node features in the input graph. \nt{If the nodes are topologically and/or feature-wise similar, we are able to verify that the GNN has successfully clustered them in the latent space.}
Conversely, one can explore how close two nodes sharing many topological neighbors are represented in the node embedding.
To achieve these goals, we present three contributions.

Our first contribution, presented in Sec.~\ref{sec:task-abs}, is a data and task abstraction for visually exploring correspondences between an input graph and the derived node embedding, to understand if a GNN model has learned important characteristics of the input graph to construct the embedding.
The abstraction is the result of extensive iterative refinement. 

To instantiate the abstraction, our second contribution is the design and implementation of an interactive multi-view interface, CorGIE (\underline{Cor}responding a \underline{G}raph to \underline{I}ts \underline{E}mbedding).
CorGIE is agnostic to specific GNN models,
with a ``grey-box'' approach that ignores the internal model details like neurons and weights and only assumes that the GNN aggregates information from neighboring nodes.  We describe the design of CorGIE in Sec.~\ref{sec:corgie-design}.

To fulfill the important task of showing topological neighbors for nodes of interest, our third contribution is a new visualization  technique, the K-hop graph layout.
The K-hop layout reveals node neighbors hop by hop, similarly to how a GNN aggregates information, and reveals the clustering structures within the node neighborhoods. Previous graph representation and layout algorithms do not suffice for this scenario.  We present the details of this technique in Sec.~\ref{sec:k-hop-layout}.


We validate our claims in two ways. We provide two usage scenarios that walk through how the CorGIE design supports the task and data abstractions in detail. We also discuss the results of a study with \nt{five} GNN experts that provide preliminary evidence of the utility and usability of CorGIE. 

\section{Background}

We first provide the minimum necessary technical background on GNNs to make this work self-contained. We then describe two concrete usage scenarios that motivate CorGIE.
We defer our discussion of the related work to Sec.~\ref{sec:related-work}.

\subsection{Graph neural networks}\label{sec:gnn-bg}
The widespread encoder-decoder perspective is a useful starting point for understanding general graph neural networks~\cite{Chami2020Machine}. The encoder combines topology, and optionally node features, to produce a node embedding. The decoder takes the node embedding to make inferences in downstream ML applications. In the GNN literature, the node embedding is an intermediate representation within the pipeline that is not necessarily of direct interest outside it.  Our perspective is different: we emphasize inspection of the learned embedding as the key to understanding. Moreover, 
the downstream ML applications are usually considered as an intrinsic component of the GNN pipeline. In contrast, we separate the downstream ML applications from the preceding pipeline so that
CorGIE can support multiple downstream applications.  CorGIE's agnostic approach supports both node classification that predicts properties of single nodes, as used for movie tagging or document topic labeling), and link prediction that predicts properties of node pairs, as used for movie recommendation. 

A GNN is usually trained using either node labels or node connections depending on the application.
For node labels, training can be supervised by separating the nodes into training, validation, and test sets if all node labels are available, or semi-supervised if only a small number of labels are available. For node connections, training \nt{can be supervised or} unsupervised. Two nodes connected by an edge are considered a positive node pair, whereas a negative node pair has no edge between them; both positive and negative pairs are sampled during training. 

A GNN learns from the input graph with \textbf{neighborhood aggregation} (aka message passing), where the embedding of a node is computed by traversing to its topological neighbors to collect and aggregate node features from them.
A GNN has multiple layers, allowing aggregation to take place hop by hop: 
each node aggregates the output of its neighbors from one layer to the next.
The input layer takes the node features, the last layer outputs a high dimensional node embedding, and the \(k\)-th layer aggregates from the \(k\)-th hop neighbor.
\nt{The example above Fig.~\ref{fig:gnn}a shows how node A aggregates information from its two hops of neighbors.}
The number of layers \(K\) is usually small, with \(K \leq 3\), indicating that it is a shallow network.
GNNs differ in the aggregation functions, ranging from a simple mean to complicated non-linear functions~\cite{peng2017cross}.
\nt{Our approach does take \(K\) as a parameter, but is fully agnostic to the details of the aggregation function.}

\subsection{Motivating scenarios}\label{sec:mot-scen}

In this paper, we refer to the target users of CorGIE as \textit{GNN developers}, 
ranging from users who consume pre-trained GNNs to experienced researchers who can propose new model architectures.

Before diving into the details of our approach and tool, we present two representative scenarios of a hypothetical GNN developer, Alice.  For each, we introduce a dataset, which consists of an input graph and the embedding constructed by the trained GNN model, and a set of visualization tasks that apply to it. 
We provide extensive additional examples of concrete tasks in Supplemental Sec.~S1. 

\subsubsection{Movie: recommendation}
Alice has a bipartite movie-user graph\footnote{Extracted from a Kaggle dataset: \url{https://www.kaggle.com/rounakbanik/the-movies-dataset}}, where an edge indicates that a user has watched or rated a movie. 
Node features in this graph depend on the node type: the movie node features are budget, popularity, vote average, \#cast, \#crew; the user node features are vote average and \#votes.
The goal is to recommend movies to users; that is, the downstream ML application is link prediction.  
Alice trains a GCN (graph convolutional network, a well-known GNN model~\cite{kipf2016semi}) with the node connections and node features in the bipartite graph, to produce a 16-dimensional node embedding.  She chooses the hyperparameter setting of 16 based on her previous experience.

To evaluate the training results, Alice wants to understand what the GNN has learned. Besides the typical quantitative metrics like accuracy, she wants to examine: 
\begin{enumerate}
    \item The overall clustering structure of the node embedding, to answer questions like \textit{Are the movies separated from the users? Are there clusters of users?}
 \item Within a user cluster, she is interested in why nodes are grouped together: \textit{Do they watch similar movies; that is, do the movies share first hop neighbors? Do they rate movies with similar scores?  Do movies they have watched have similar budgets?}
 Similarity in topology and/or feature values indicate the GNN has done a good job in grouping these users.
 \item Between two user clusters, she wants to see their differences: \textit{Do they watch different sets of movies? Do they rate movies differently?} Significant differences indicate that the GNN has done a good job in differentiating the two user groups.
 \item Instance-level inspection of movie recommendations is Alice's next undertaking. She wants to spot-check the list of recommended movies for some user, for example a recommendation of \textit{Interstellar} for a viewer who has seen \textit{The Matrix} and \textit{Guardians of the Galaxy}.
 She wants to understand why the GNN has generated them: \textit{Are recommendations determined by shared viewing patterns, as with collaborative filtering~\cite{ricci2011introduction}? Are they motivated by similar node features? Do they capture topic similarity, like science fiction?} 
\end{enumerate}

\subsubsection{Cora: paper topic labeling}
The Cora dataset is commonly used as a benchmark in the GNN literature~\cite{mccallum2000automating}. It is an academic paper citation graph, where the node features are a dictionary of 1433 one-hot vectors showing whether specific words exist in papers.  The downstream ML application is to classify the papers according to a set of given topics (classes).  Alice trains a GAT (graph attention network, another well-known GNN model~\cite{velivckovic2017graph}) with the node labels in the Cora graph, and produces a 7-dimension node embedding \nt{representing a probability distribution over the} 7 classes of paper topics. She thus converts the values for each dimension to probabilities of belonging to that class, which is a common practice in the GNN literature.

Following a similar analysis process for the movie scenario, Alice first explores the clustering structure by visualizing shared topological neighbors and similarity of node features from an overview level. 
Then, she tries to understand the pattern of misclassified nodes.  She selects groups of these nodes that are either located in the same area of latent space or have the same predicted label; that is, the same error. She inspects the shared neighbors and the words within groups to determine whether there are commonalities to blame for the misclassification.

\section{Data and task abstraction}

After considering many examples of concrete tasks faced by GNN developers, we generalized them to understand the problem at a higher level. We formed the data and task abstractions iteratively, over several months, through four parallel thrusts: review of the GNN and visual analytics literatures, informal interviews of four GNN developers within three different organizations, development and use of prototypes of the CorGIE interface, and reflection between paper authors.

\subsection{Overview}\label{sec:task-overview}

We generalize the many concerns faced by GNN developers as they train their models into three driving questions. For memorability, we frame them as questions that might be asked on a road trip: 
\begin{itemize}
    \item Q1: \textbf{are we there yet:} should we train or tune more?
    \item Q2: \textbf{are we lost:} does it behave as we expect?
    \item Q3: \textbf{what's that:} what does this exemplar do? 
\end{itemize}

The first big-picture question concerns the potential need to keep training the model for more epochs, tuning the hyperparameters, 
or reconsider how they construct the graph. It would be answered by understanding whether the results are satisfying. 
The second question, also big-picture, concerns developers' understanding and trust in whether the GNN learns inherent characteristics from the input graph; that is, does it behave as they expect. These two big-picture questions are particularly difficult to answer with previous evaluation approaches, which rely heavily on quantitative metrics for the downstream ML application; we aim to provide visual and qualitative evidence and insights to answer these two questions in a more detailed and thorough way. The third question is more specific, concerning investigating exemplar instances of nodes or links. 
Some previous approaches like the GNNExplainer~\cite{ying2019gnnexplainer} do support such single-instance inspections, but we seek to make them easier and faster. 

\begin{figure*}[!t]
  \centering
  \includegraphics[width=.8\linewidth]{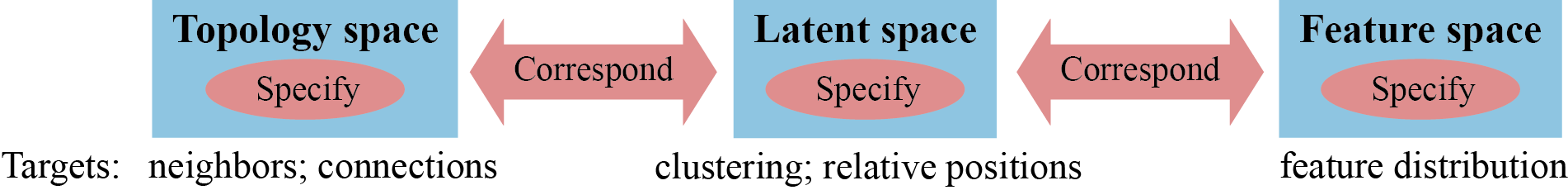}
  \caption[Data and task abstraction]{Data and task abstraction. Users can specify items within all three data spaces, and investigate correspondences between the latent space and topology or feature spaces. The targets of activity in each space are listed below it.} 
  \label{fig:task}
\end{figure*}

We posit that finding the correspondences among 1) the node positions and clustering structure in the latent space, 2) the topological neighbors in the input graph, and 3) the node feature distributions in the input graph, can shed light on these questions. 
We thus discuss three abstract data spaces, as illustrated in Fig.~\ref{fig:task}: latent space, topology space, and feature space.

\subsection{Data Space Abstraction}\label{sec:data-spaces}
Our data abstraction features three sub-spaces: the latent space, the topology space, and the feature space. We compute distance metrics in each of these spaces. We also incorporate prediction results if available. 
We follow the order of views in the CorGIE interface to introduce the three spaces.

\subsubsection{Latent space}
The \textbf{latent space} is where the node embedding learned by the GNN lives. Each node in the input graph has its own high-dimensional vector representation. 
The number of dimensions is usually less than a thousand, although this number does not affect the complexity of the tasks and thus is not a constraint for the CorGIE design.

In the latent space, the absolute vector values are usually not directly interpretable. We are interested in the vector similarities and particularly, whether there is a clustering structure among the node vectors.
For example, in the \textit{Cora} scenario, a relevant question is whether there are exactly seven clusters of papers in the latent space, which would match the seven topics in the ground truth.

The scope of this paper is node embedding; less popular cases like edge embedding and graph embedding are left for future work. 

\subsubsection{Topology space}
The \textbf{topology space} consists of the topological connectivity of one single input graph.
The input graph could be either homogeneous, with a single type of nodes, or heterogeneous, with multiple node types. The design target for CorGIE is to handle up to 6 node types, which covers many useful cases for heterogeneous graphs~\cite{yang2020heterogeneous}. For graph size, CorGIE will accommodate up to 20K nodes, to handle many prevalent benchmark datasets in the GNN literature~\cite{dwivedi2020benchmarking}.

In the topology space, we are interested in the topological neighborhoods and the connections (edges) between node pairs. GNNs aggregate information along edges to compute the node embedding. We thus derive node's \textbf{neighbor set} containing its k-hop neighbors; that is, the set of all of the nodes reachable from a node \textit{v} within \textit{k} topological hops.

We limit our scope to one input graph, leaving the multiple graph case to future work. We also leave the support of multiple edge types for the future.

\subsubsection{Feature space}\label{sec:feature-space}
The \textbf{feature space} consists of all features (often referred to as attributes in the visualization literature) of the graph nodes.
We distinguish between dense and sparse features.
Dense features are numeric or boolean features that can be collected and understood independently, with interpretable semantic meanings for each one, such as \textit{budget} or \textit{average vote} from the \textit{Movie} scenario. 

Sparse features are usually collected together and share a combined semantics, where interpretation typically takes place across the entire set considered as a whole. Typically they are stored as
one-hot encoded categorical features like city names or the dictionary of words from the \textit{Cora} scenario, where the collection is represented by hundreds or thousands of bit vectors. 

In the feature space, we are interested in the distributions of feature values, which involves deriving aggregations.  
We deep-dive into the details of feature aggregation in Sec.~\ref{sec:node-feature-view}.

The design target of CorGIE is to handle up to a dozen dense features and up to 2K sparse features. 
Note that we deem edge or graph features out of the scope of this paper.

\subsubsection{Distance metrics}\label{sec:distance-metrics-abs}
To quantify correspondences between spaces, we introduce a distance metric between any pair of node within each space, so that scalar distance values across different spaces can be compared. 
In the latent space, we derive the cosine distance of the embedding vectors, which is commonly used in the downstream ML applications.
In the topology space, we derive a topologically oriented distance measure, the inverse Jaccard index for the full set of k-hop neighbors of each pair of nodes in the input graph.  This measure is 
based on the Jaccard Index: the intersection over union of the neighbor set. It is intuitive and easy to compute, but the flattened neighbor sets are sometimes an oversimplification because neighbors in different hops cannot be distinguished from each other. 
In the feature space, we derive the Euclidean distance of feature vectors (scaled linearly to \([0,1]\)), which is also familiar to GNN developers.

\subsubsection{Prediction results (optional)}
After training is complete, the GNN can be used for prediction in the downstream ML applications (Fig.~\ref{fig:gnn}).
We leverage prediction data by comparing the predicted results to ground truth when available, to derive true/false negatives and positives. 
For node classification applications, we compare the true and predicted node labels to derive a label correctness value (\textit{correct} vs \textit{wrong}) for each node. For link prediction applications, we obtain the predicted positive and negative node pairs, and compare these to the edges of the input graph.  We derive labels for two interesting categories of node pairs: \textit{false positives}, where unconnected node pairs are predicted as connected (recommended pairs); and \textit{false negatives} which are node pairs connected with an edge but predicted as disconnected.  We also provide options to derive the true positives and negatives if desired.
Although the fundamental data and task abstractions do not rely on these prediction results, they can be used for filtering. 

\subsection{Task abstraction}\label{sec:task-abs}

We identify a unifying task abstraction built around an iterative cycle of two phases, as shown in Fig.~\ref{fig:task}:
\textbf{specify} items in any of the three data spaces, and \textbf{correspond} items between either the latent space and topology space, or the latent space and feature space. Notably, our task abstraction does not entail finding correspondences between the topology and feature spaces, because that exploration would focus only on characteristics within input graph and would not provide direct insight into GNN behaviors. 

The \textit{specify} step allows users to indicate nodes of interest in any of the spaces, based on the targets visible within each of them. 
In the latent space, they could specify nodes based on relative positions with respect to any visible cluster structure, or with respect to the latent distance distribution.
In the topology space, they could specify nodes based on neighbors, cliques, or the topological distance distribution.
In the feature space, they could specify nodes of specific feature values. 
If available, they can also specify a collection of nodes based on the prediction results such as label correctness or newly predicted links between nodes.
Below, we refer to a group of specified nodes as a focal group, which has a double meaning: it is the mental focus within a user's thoughts at the current step of their data analysis process, and it is the focus of actions and computation within the CorGIE interface.

The \textit{correspond} step then allows users to explore to what extent the characteristics of the specified nodes in one space
correspond to those in the other spaces.
For example, for a tight cluster in the latent space, 
users could verify whether the nodes in the cluster
share many neighbors in the topology space and have similar features in the feature space.
If so, the GNN has successfully learned how to group these nodes together. The extent to which these spaces line up with each other is a qualitative judgement call, not a precisely computable metric. The intent of this abstraction is to support GNN developers in obtaining higher confidence and trust in their models than would be possible from purely quantitative summary metrics such as accuracy. 

The abstraction supports the process of \textit{specify} and \textit{correspond} as an iterative loop, where exploring correspondences can trigger the user's interest in specifying other sets of nodes.  
For instance, the user might start by specifying a cluster in the latent space, and after checking its correspondences to the topology space would become curious to then specify a sub-cluster within it to explore further connections. Multiple cycles of refining and changing the sets of specified nodes based on the visual correspondences allow users to connect the dots and find answers all the three of the driving questions. 

This task abstraction encompasses all three of our driving questions. It was developed through considering the commonalities between all of the concrete tasks that we analyzed, not only those presented in the motivating scenarios (Sec.~\ref{sec:mot-scen}) but also many additional concrete examples summarized in Supplemental Sec.~S1. During the abstraction process, we initially considered a much more complex set of targets and actions, such as analyzing outliers with respect to a cluster, or analyzing one node vs.~all other nodes, or whether verifying differences should be treated differently than verifying similarities. In the end, we realized that almost all of these questions could be framed much more simply, in terms of specifying a very small number of groups as a target and then checking on correspondences. It was rare to require even three groups: one or two groups were sufficient for almost all analysis questions. Our final design is optimized for up to two groups, although it is possible to specify several more to handle edge cases.

\section{CorGIE design}\label{sec:corgie-design}
Based on the data space and task abstraction, we design CorGIE, a multi-view tool that reveals the correspondence between an input graph and its embedding.  We describe the design of each view, then the view coordination between them, and finally the implementation.

\begin{figure*}[!t]
  \centering
  \includegraphics[width=\linewidth]{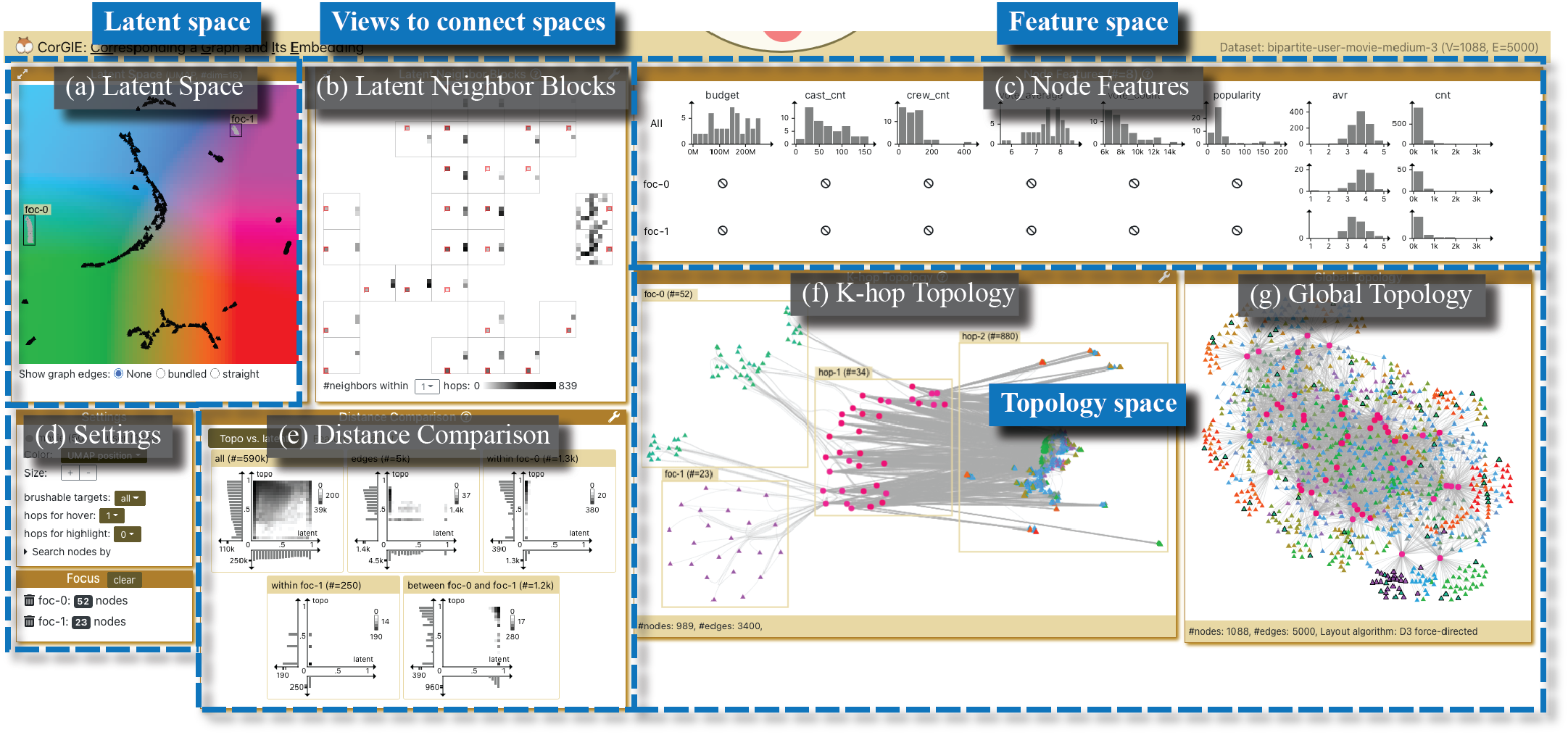}
  \caption[Full screenshot of CorGIE on the Movie dataset]{Full screenshot of CorGIE interface on the \textit{Movie} dataset, with two focal groups of user nodes.
    The views, with names shown in grey boxes, are laid out in four major areas on the screen, shown in blue: (a) the \viewname{latent space view} is for 2D node positions in latent space; (c) the \viewname{node features view} is for feature distribution of all and focal nodes; (f \& g) the \viewname{topology views} show local topological neighbors and global topology; (b) the \viewname{latent neighbor blocks view} and the (e) \viewname{distance comparison view} connect different spaces. The fifth area is for toggles and menus: (d) the \viewname{settings view}.
  }
  \label{fig:teaser}
\end{figure*}

\subsection{Overview}
Fig.~\ref{fig:teaser} shows an overview of the CorGIE interface on the movie scenario, which has seven main views. 
\nt{Using the language in Sec.~\ref{sec:data-spaces} and Fig.~\ref{fig:task}, a view either shows targets in a specific space or connects multiple spaces.}
The \viewname{latent space view} is on the top left. Two views at the bottom right show the topology space, the
\viewname{global topology view} and a novel graph layout for a user-chosen subset of nodes in the \viewname{K-hop topology view}. The feature space is showcased at the top right in \viewname{node features view}. To enable direct and quick correspondence exploration, we have two views in the middle that can incorporate two or three spaces simultaneously. We introduce the \viewname{latent neighbor blocks view} to connect the latent and topology space, and the \viewname{distance comparison view} to connect all three spaces. Finally, there is a \viewname{Settings} view at the bottom left.

To support the two steps of \textit{specify} and \textit{correspond} in the task abstraction, the interaction design of CorGIE involves three actions: \textit{hover}, \textit{select}, and \textit{focus}.  

Node specification occurs through a \textit{focus} action, and correspondence exploration is triggered through \textit{hover} and \textit{select} actions. The \textit{focus} action creates one or a few special node groups of interest, denoted by \textit{foc-0}, \textit{foc-1}, and so on.
When \textit{hovered} or \textit{selected}, nodes are then highlighted across views.  When nodes are \textit{focused}, there will be a novel graph layout showing their k-hop topological neighbors.
\nt{The interface as a whole addresses the three questions in Sec.~\ref{sec:task-overview} using all the views and interactions.}

\subsection{View design}
We introduce the visual encoding and design rationale for each view.

\subsubsection{Latent space}
The \viewname{latent space view} on the upper left of the interface (Fig.~\ref{fig:teaser}a, Fig.~\ref{fig:movie-check-overview}a, Fig.~\ref{fig:cora-overview}a) shows the clustering structure and the relative positions of nodes in the latent space \nt{(targets in Fig.~\ref{fig:task})}. We use UMAP~\cite{mcinnes2018umap} to project the latent space to two dimensions and plot the dimensionally reduced nodes as a scatterplot.  
GNN developers are typically familiar with dimensionality reduction and are thus aware that loss of information is inevitable in this process. 

CorGIE supports color coding nodes by their UMAP positions in all views, and 
in that mode a rainbow background for the entire 2D latent space view provides a salient reference for positions within the latent space, as shown in 
Fig.~\ref{fig:teaser}. We chose a highly saturated CIELAB colormap (x axis for the variable \(a\), and y axis for \(b\)) that has substantial color exploitation and strong performance for the synoptic localization task of detecting groups of nodes~\cite{bernard2015survey}. The combination of dimensionality reduction and this colormap shows latent space distances, so that similar nodes receive similar colors whereas dissimilar nodes receive dissimilar colors.

Nodes are shape-coded for heterogenous graphs with multiple node types, with the same shape preserved across the latent and topology space views. Edges are off by default to avoid clutter but can be shown on demand in this view, as straight or bundled edges.

\subsubsection{Global topology}
We present graph topology with node-link diagrams in CorGIE.
We adopt the D3 force-directed layout~\cite{bostock2011d3} for the \viewname{global topology view} (Fig.~\ref{fig:teaser}g, Fig.~\ref{fig:cora-overview}b). It visualizes the topology of an entire input graph, and is computed once in the pre-processing step.
As in the \viewname{latent space view}, we encode graph nodes as circles for homogenous graphs and glyphs of different shapes for different node types for heterogeneous graphs. We use straight lines for graph edges and curved lines when edge bundling is enabled.

Although this straightforward backstop layout may provide useful insights in smaller graphs, such global views can be extremely cluttered for larger ones. In that case, our custom layout for exploring local neighborhoods of user-specified subsets of the graph is crucial, in the k-hop topology view.  

\subsubsection{K-hop topology}

The \viewname{K-hop topology view} (Fig.~\ref{fig:teaser}f, Fig.~\ref{fig:movie-recomm}b \& e, Fig.~\ref{fig:cora-overview}d, Fig.~\ref{fig:cora-buggy-pair}b, Fig.~\ref{fig:cora-case-study}d \& e) shows the topological neighbors of the focal node sets specified by users \nt{(targets in Fig.~\ref{fig:task})}.  We devise the k-hop layout algorithm to mimic how information is aggregated in GNNs, with boxes enclosing meaningful groups. On the far left are the focal nodes, with hop-1 neighbors in the middle, hop-2 neighbors to the right, and hop-3 neighbors on the far right. 
Within the boxes, nodes are clustered using the topological (Jaccard) distance.  Full algorithmic details on the layout are provided in Sec.~\ref{sec:k-hop-layout}.

\subsubsection{Node features}\label{sec:node-feature-view}

The \viewname{node features view} on the top right (Fig.~\ref{fig:teaser}c) shows feature distributions \nt{(targets in Fig.~\ref{fig:task})} for up to three sets of nodes: all nodes, and each of the two possible sets of focal nodes (\textit{foc-0} and \textit{foc-1}).  Dense and sparse features are visualized differently due to their scale differences.

For dense features, we choose histograms as they are intuitive and well-known for visualizing and comparing distributions of scalar values. They also scale well for the available display space, to handle the design target of up to a dozen dense attributes. 
We organize these feature histograms as a matrix.
Each column represents one feature, whereas the rows show different collections of nodes.
The first row shows the full distribution of all nodes; the next row shows the distribution of the focal node set \textit{foc-0} and underneath that is the set \textit{foc-1}. 
For example, in Fig.~\ref{fig:teaser}c, the two focal groups of user nodes have different distribution in average vote (the second last \textit{avr} column), but little difference in the number of votes (the last column).
Note that for heterogeneous graphs, different types of nodes can have different features, so focal node sets could end up with some empty histograms.  \nt{The two focal groups in Fig.~\ref{fig:teaser} contain user nodes only, and thus there is no data for the six movie attributes.}

In contrast to dense features, there could be thousands of sparse features, and their ordering is not explicitly meaningful. 
\nt{Therefore, the visual encoding for sparse features is different from the histograms for dense features, and we include a full screenshot in Supplemental Fig.~S14 showing sparse features for readers to compare Fig.~\ref{fig:teaser} showing dense features.}
We designed a two-level custom view to make effective use of the screen space in a compact area, as shown in Fig.~\ref{fig:cora-overview}f. The top part is a \textit{feature strip} overview that aggregates the information in the more detailed \textit{feature matrix} part below it. The lower \textbf{feature matrix} is a heatmap containing one square cell per feature (e.g., a word \texttt{train}), and its luminance encodes the count of nodes with that feature (e.g., the number of papers with the word \texttt{train} in the Cora dataset).  Conceptually, the heatmap contains a single very long line that is simply wrapped to fit into a rectangular aspect ratio. It is not a true 2D matrix, so the absolute position of a cell in terms of a row/column is not meaningful.  
The upper \textbf{feature strip} aggregates the information in the heatmap into a highly compressed pixel-based depiction where many consecutive features are combined into the same vertical line, with a luminance representing the maximum values across the range. The number of features to aggregate is determined by the available horizontal pixel budget of the view, so that the strip fits within it.

As with the dense features, there is a row for all nodes on top, with a row for each of the two focal groups below to show partial feature distributions. In addition, there is a lower \textit{diff} row derived by subtracting the feature values of the two focal node groups. Fig.~\ref{fig:cora-overview}f shows an example (bottom row) that visualizes differences in the word distribution of \textit{foc-0} and \textit{foc-1}, where the dark strips indicate that there are large difference in those word counts. 

In each row, the \textit{feature matrix} can be hidden to save space, or expanded to inspect detail. The luminance values are separately normalized within each row since the value ranges could be quite different, so they each require a separate legend.
 
\subsubsection{Latent neighbor blocks}

The \viewname{latent neighbor blocks view} (Fig.~\ref{fig:teaser}b, Fig.~\ref{fig:cora-overview}c) overlays topological neighbor distributions on the 2D latent space. The challenge is to show the positions of each node's neighbors in the latent space, for each of the nodes. We do so with aggregation and nesting. We aggregate by partitioning the space into an 8\(\times\)8 grid, to create a coarser representation of the space with 64 \textit{blocks}, and map each node as belonging to the block that encloses it. Within each block, we nest a complete copy of the latent space itself, again at a coarse resolution as an 8\(\times\)8 grid of \textit{cells}. Blocks that do not contain any nodes are omitted from the drawing. 

This view is inspired by 
origin-destination (OD) maps for geo-spatial networks~\cite{wood2010visualisation}, but we show neighbor set distributions rather than geographical movement. 
Within each high-level block, we outline the single low-level cell that corresponds to its block index with red, like the \textit{origin} in an OD map.
For every other cell in the block, we encode with luminance the number of neighbors of nodes that fall within the red-outlined origin cell, like the \textit{destinations} in an OD map. 
If the red-outlined block and its surrounding cells are darker than others in all blocks, that pattern indicates neighbors in the topology space are still neighbors in the latent space, and the GNN has successfully preserved neighborhood structure.  Fig.~\ref{fig:cora-overview}c for the \textit{Cora} scenario illustrates this case. 

However, there is a limitation to this visual encoding: it is less informative for a bipartite graph, such as the \textit{Movie} scenario, where movie nodes are only allowed to have user nodes as their first hop neighbors. In this case, \nt{Fig.~\ref{fig:teaser}b does not offer much insight}.

\subsubsection{Distance comparison}

The \viewname{distance comparison view} (Fig.~\ref{fig:teaser}e, Fig.~\ref{fig:movie-recomm}e \& ~\ref{fig:movie-recomm}f) shows distributions of distances in each space and supports comparison between spaces.  In the data abstraction (Sec.~\ref{sec:distance-metrics-abs}), we derive one distance metric in each space. This view reveals matches and mis-matches of distances between the topology and the latent space in one tab, and between the feature and the latent space in a second tab.
A match constitutes a positive correlation of the distances in two spaces. This view shows pairs of nodes, which may either represent an edge or be disconnected in the input graph. 

To present and compare two scalar distributions simultaneously, we combine two histograms and a gridded scatterplot into one chart, where the x-axis represents the distance distribution in the latent space, and the y-axis represents that in the topology or feature space. The number of items to plot in the scatterplots can be huge, so we use a grid-binning approach to avoid over-drawing, and to speed up computation we also down-sample the node pairs if there are more than a million. 

Each tab accommodates multiple charts to show different sets of node pairs side by side, such as all, within a focal group, between two focal groups, or a user-customized filtered set based on connectivity and link prediction values (when available). The customization options are shown in Supplemental Fig.~S18, including a choice between linear and log scale to handle the variance in data.

For example, the three bottom histograms in Fig.~\ref{fig:movie-check-overview}e show the latent distance distribution within \textit{foc-0}, within \textit{foc-1}, and between the two groups respectively.  We observe that the between-group distances are much larger than the within-group distances, indicating that the two focal groups are two distant clusters in the latent space.
Fig.~\ref{fig:cora-buggy-pair}a shows a pursuit of problematic node pairs, where
brushing and highlighting the bottom-right area in the scatterplot signals a mis-match with a negative correlation between latent and topology distance.

\subsection{View coordination}
To visually underscore the linkages between the views, we carefully maintain the consistency of visual encodings for nodes across views including the shape, color, and size.
The node color is user configurable in the \viewname{Settings} view (Fig.~\ref{fig:teaser}d), to 
either the latent space position (the similarity rainbow), a specific node feature (a sequential ramp), node type (distinguishable categorical colors), or node classification labels if available (also categorical). The shape depends on the node type for heterogeneous graphs, or is a circle in the homogeneous case. The size is also globally configurable in \viewname{Settings}. 

We also develop user interactions with consistent semantics across all views.
In CorGIE, there are three major interactions from light weight to heavy weight: \textit{hover}, \textit{select}, and \textit{focus}.
Typically, they are used in order: \textit{hover} is for quick and temporary exploration, then \textit{select} offers a more persistent visual prompt for nodes of interest that are identified with hover, and finally \textit{focus} stores the currently selected nodes into a persistent group. 
The look and feel of the interaction is shown in the supplemental video.

The \textit{hover} action is triggered when users mouse over objects, such as a node or edge in the \viewname{topology views}, and a block in the \viewname{latent neighbor blocks view}.
On \textit{hover}, CorGIE reacts with strong visual prompts (Fig.~\ref{fig:movie-recomm}c \& e): strokes on activated nodes and edges while desaturating the background with a half-transparent mask, black partial distributions on top of the node feature histograms, tooltips and node labels.

The \textit{select} action is triggered by clicking a node/edge or brushing multiple nodes.  The visual prompt is highly similar to that for \textit{hover} (Fig.~\ref{fig:movie-check-overview}c \& d), but persists even when users move the cursor away. Users can control whether the neighbors of target nodes are highlighted during the \textit{hover} and \textit{select} actions, with drop-down menus in the \viewname{Settings View}.

The heavy-weight \textit{focus} action enables users to find correspondences for one or two groups of specified nodes. When a focal group is created or modified many views are updated, with changes to the \viewname{features} and \viewname{distance} views and the computationally intensive operation of creating a new \viewname{K-hop topology} layout. CorGIE supports several focus actions: create a new focal group, add-to/single-out/remove-from existing groups, and clear groups. These actions enable users refine and change the specified nodes, as required by the iterative nature of the tasks (Sec.~\ref{sec:task-abs}). 
Since only selected nodes can be the target of focus actions, the focus menu to choose one of these actions appears only after nodes are selected. It drops down from the top of the window and looks like a corgi dog's paw, in keeping with the system name, as shown in the video and Supplemental Fig.~S6. 
The focal nodes are bounded by boxes and labelled with \textit{foc-0} or \textit{foc-1} in both the \viewname{focal layout} and \viewname{latent space} views.

\subsection{Implementation}
CorGIE is implemented using ReactJS~\footnote{\url{https://reactjs.org/}} and Redux~\footnote{\url{https://redux.js.org/}} as the frontend scaffold.
We choose Canvas over SVG in most of the views for browser performance.  For a graph with more than a thousand nodes, there would be thousands of DOM elements such that re-rendering SVGs on user interactions would be very expensive.  We develop a layering system using the Konva library~\footnote{\url{https://konvajs.org/}} to add visual elements on top of a static canvas in order to avoid the expensive re-render whenever possible.
As the \textit{focus} action involves heavy computation that takes seconds and even minutes, we offload it to multiple Web Workers to keep the application responsive, and use the Comlink library~\footnote{\url{https://github.com/GoogleChromeLabs/comlink}} to communicate between threads.

\subsection{Design alternatives}

During the iterative refinement process for the CorGIE interface, we experimented extensively with different alternatives for exploring the neighbor sets of a single focus group, including an Upset-based~\cite{lex2014upset} view for neighbor sets and a partial adjacency matrix with roll-up histograms. We discuss these design alternatives in detail in Supplemental Sec. S3.
However, the pixel space required would be infeasible for realistic sizes of graphs (e.g. graphs with hundreds of nodes), and there was no obvious extension to comparing neighbor sets between two focus groups. Our final design relies more extensively on interactive exploration to explore neighbor sets for one or two groups at a time, rather than attempting to visually encode all of possibilities simultaneously. 
In the case study sessions, a domain expert suggested that we could use the traditional bipartite layout for the \textit{Movie} scenario, but we consider the force-directed layout a more general solution for all kinds of graphs.

\section{K-hop layout}\label{sec:k-hop-layout}
We present a new visual encoding technique, the K-hop graph layout, which is used in CorGIE's \viewname{K-hop topology view} (Fig.~\ref{fig:teaser}f, Fig.~\ref{fig:movie-recomm}b \& e, Fig.~\ref{fig:cora-overview}d, Fig.~\ref{fig:cora-buggy-pair}b, Fig.~\ref{fig:cora-case-study}d \& e) \nt{to show neighbors in hops for nodes of interest.
This encoding aligns with the concept of neighborhood aggregation in GNN, and reflects how users think about the model.}
We present the computational scalability of the K-hop layout in this section, and evaluate its suitability for the tasks in Sec.~\ref{sec:results}.
We also discuss three design alternatives that we considered and their limitations. 

\begin{figure*}[!t]
  \centering
  \includegraphics[width=\linewidth]{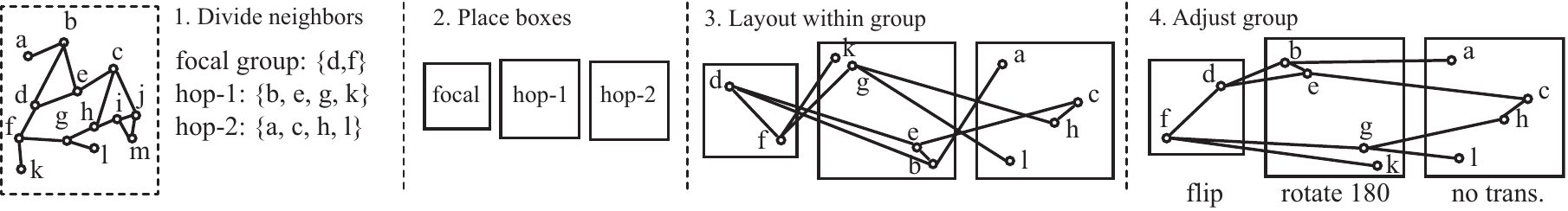}
  \caption[K-hop layout algorithm in four steps]{K-hop layout algorithm in four steps. The optional fifth step is bundle edges (not shown).}
  \label{fig:k-hop-layout}
\end{figure*}

\subsection{Algorithm}
We first provide an overview of the layout algorithm, then the technical detail and rationale of each step, and finally we discuss its scalability.

The K-hop layout aims to organize topological neighbors of user-specified nodes by the number of hops and their clustering structure.
As illustrated in Fig.~\ref{fig:k-hop-layout}, it is computed in four main steps: 1) divide relevant nodes into groups, 2) bound nodes within boxes and lay out the boxes, 3) lay out nodes within each box independently, and 4) perform transformations to optimize global readability. An optional final step is 5) edge bundling.
\nt{We follow this top-down layout strategy to ensure that the most important information is displayed with an effective visual channel: spatial separation.}

\textbf{Divide neighbors.}  We divide the nodes into groups: user-specified focal groups, all 1st-hop neighbors of focal nodes, all 2nd-hop neighbors of focal nodes, and so on until the Kth hop (where K$\leq$3).  The rest of nodes are deemed irrelevant and thus discarded, so this view shows only a subset of the nodes in the global layout. This division naturally matches the neighborhood aggregation in GNNs (Sec.~\ref{sec:gnn-bg}), by design.
Note that a node can only appear in one group to avoid confusing duplication, with priority given to the leftmost view in which it appears.

\textbf{Place boxes.}  We bound the nodes of groups within boxes and place the boxes from left to right. For the focal groups on the left, multiple focal nodes are placed from top to bottom. This layout mimics how the information flow is typically thought about by GNN developers, as we verified in our interviews with them. 

\textbf{Lay out within group.}  We lay out the nodes within each group independently.
To reveal the clustering structure of nodes, we use a dimensionality reduction technique  UMAP~\cite{mcinnes2018umap} to reduce a high-dimensional topological distance matrix of nodes down to two dimensions.  Users can choose between local distance (connections to previous hop only) or global distance (all K hop connections).
The layout falls back to D3 force-directed layout if there are not enough nodes to run UMAP.  Because of the independence between groups, we can parallelize the UMAP processes for performance improvement.  It would be possible to use alternative layout algorithms in this step like t-SNE~\cite{van2008visualizing}; we choose UMAP for its speed and strength in revealing global structure~\cite{coenen2019understanding}.

\textbf{Adjust group.}  We need to fine-tune the layout from previous step since purely local layout decisions could endanger the global readability.
For example, there can be many edge-edge crossings in step 3 of Fig.~\ref{fig:k-hop-layout}, even though the local layout within each box has been optimized. We use an approach similar to Procrustes analysis~\cite{borg2005modern} to potentially reorient each group.
Specifically, we can apply six possible transformations to each group: rotation (0, 90, 180, 270 degrees), and flip (horizontal and vertical). These rigid transformations do not change the relative positions within the group.
We find the optimal transformations for each group with a simple enumeration algorithm, that is, to enumerate all 6 possible transformations for all groups for a total of \(6^B\) combinations, where \(B\) is the number of groups.
The node-repulsion Linlog function~\cite{noack2007energy} is used as the readability metric to find the optimum. 
Note that only node pairs between boxes are considered in the measurement, and we sample them randomly to compute an approximation of the function for performance speedup.

\textbf{Bundle edges.} Finally, to further reveal the pattern of connections, we apply edge bundling to the graph layout to reduce visual clutter.  Out of several edge bundling algorithms~\cite{zhou2013edge}, we choose the multilevel agglomerative one~\cite{gansner2011multilevel} for its speed and competitive visual performance. We note that it can introduce distorted edges, which are harder to trace than to unchanged ones, so we allow the user to toggle between straight and bundled edges. 

\subsection{Scalability}

\begin{table}[]
  \scriptsize
  \begin{tabular}{rrrrrr|rr}
  \multicolumn{1}{l}{} & \multicolumn{1}{l}{} & \multicolumn{1}{l}{} & \multicolumn{1}{l}{} & \multicolumn{1}{l}{} & \multicolumn{1}{l|}{} &
  \multicolumn{2}{c}{Times (sec)} \\ \hline
  \multicolumn{1}{c}{Dataset} &
  \multicolumn{1}{c}{\begin{tabular}[c]{@{}c@{}}N/E \\ total\end{tabular}} &
  \multicolumn{1}{c}{\begin{tabular}[c]{@{}c@{}}N \\ k-hop\end{tabular}} & \multicolumn{1}{c}{\begin{tabular}[c]{@{}c@{}}E \\ k-hop\end{tabular}} & \multicolumn{1}{c}{\begin{tabular}[c]{@{}c@{}}N \\  box\end{tabular}} & \multicolumn{1}{c|}{B} & \multicolumn{1}{c}{\begin{tabular}[c]{@{}c@{}}UMAP\\ (step 3)\end{tabular}} & \multicolumn{1}{c}{total} \\ \hline
  Movie & 1K/5K & 1088 & 5000 & 633 & 4 & 4.1 & 5.5 \\
  Movie & 1K/5K & 1088 & 5000 & 370 & 5 & 2.5 & 4.7 \\
  Cora & 3K/5K & 989 & 2010 & 576 & 3 & 3.3 & 3.9 \\
  Cora & 3K/5K & 2116 & 4222 & 1028 & 4 & 8.5 & 9.6 \\
  Coauthor & 10K/54K & 9528 & 43317 & 4254 & 3 & 47.1 & 57.6 \\
  Coauthor & 10K/54K & 9951 & 44261 & 6544 & 4 & 110.1 & 123.9 \\ \hline \\
  \end{tabular}
  \caption{Computational benchmarks for the K-hop layout. N is \#nodes, E is \#edges: approximate for full graph and exact for used within k-hop layout. N box is maximum \#nodes in any box (key dependency), B is \#boxes; times in seconds.}
  \label{tab:comp-time}
\end{table}

To aid the perceptual scalability of our approach, our emphasis is on reducing edge-edge crossings, which would hinder users' judgement about topology and connectivity. Our choice to reveal clustering structure through the K-hop layout constitutes a trade-off, where there could be substantial node-node overlap in the K-hop layout.

For computational scalability, we instrument the code to obtain elapsed (wall-clock) timings, and measure the running times with different interactively chosen focal groups for three datasets: \textit{Movie} (N=1K, E=3K) and \textit{Cora} (N=3K, E=5K) as introduced in Sec.~\ref{sec:mot-scen}, and \textit{Coauthor} (N=10K, E=54K) extracted from a larger graph of 69K nodes~\cite{tang2008arnetminer}. 
The experiment is conducted in Chrome on a 2020 MacBook Pro with 2.3GHz 8-core i9 CPU.  
We present the results in Tab.~\ref{tab:comp-time}.  In addition to the total graph size, we provide the number of nodes and edges used within for the k-hop graph layout (the combination of the focal nodes and their neighbors within k hops). We see that sometimes the k-hop focal layout incorporates all nodes and edges but sometimes shows only a subset of them, depending on the choice of focal groups. 

Graphs with 1K nodes like those in the \textit{Movie} dataset take a few seconds to compute, whereas those with 10K nodes take significantly longer.  
The UMAP computation (step 3) is the computational bottleneck in our algorithm, as shown the timing numbers on the two rightmost columns.  
We note that the times do not depend on the total number of nodes, but instead, the largest number within a single bounding box, because we run UMAP processes in parallel across multiple worker threads for each box. We thus include the node count within the maximum box in the table.
For example, the time for the last run (with 6544 nodes in the most numerous box) of 110 seconds doubles that of the previous run (4254 nodes maximum) of 50 seconds on the same \textit{Coauthor} dataset. 
Compared to the UMAP computation, the time required for step 1 (divide neighbors), 2 (place boxes), and 4 (adjust group) is negligible; the optional 5th step (edge bundling) takes about \(10-20\%\) of the total time.

\subsection{Design alternatives}
We tried three design alternatives to the \viewname{K-hop topology}, which were less effective at showing neighbor hops and clustering structure than our final choice: D3 force-directed, WebCola, and space-filling spirals. 

\begin{figure*}[!t]
  \centering
  \includegraphics[width=\linewidth]{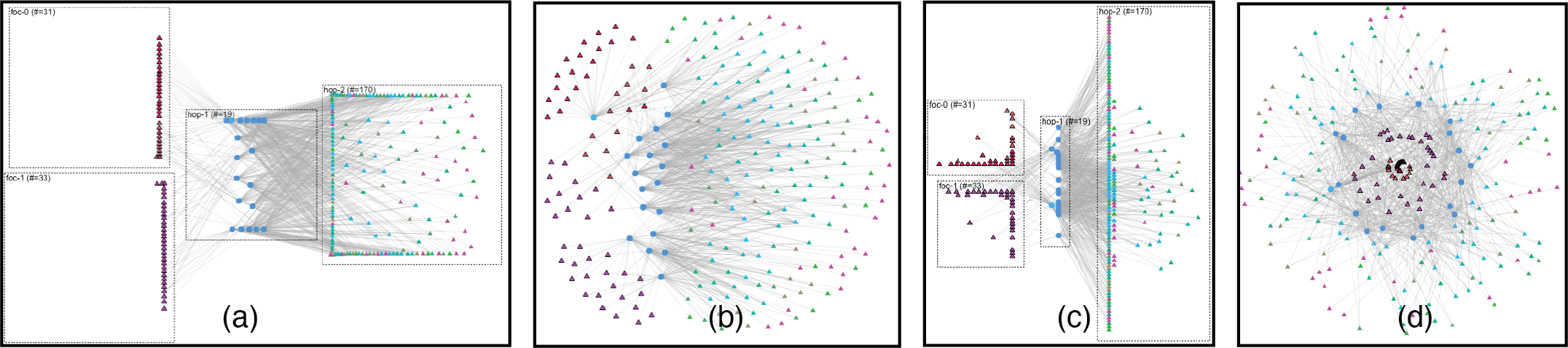}
  \caption[Design alternatives to K-hop layout]{Design alternatives to K-hop layout: (a) D3 force-directed layout with strict bounding box; (b) unbounded D3 force-directed layout with repulsion force between groups; (c) WebCola force-directed layout with constraints; (d) space-filling curve layout using spiral curve.}
  \label{fig:design-alter}
\end{figure*}

The D3 force-directed layout is versatile as we can configure the type of strength of forces.  Similarly to the K-hop layout, we use topological distance as the pulling force between nodes, instead of the usual approach of using the edges to control the forces.  Fig.~\ref{fig:design-alter} (a) and (b) shows two versions of this layout.
The first take is bounded by boxes like the K-hop layout, but the nodes are pulled towards other groups and positioned alongside the group boundary as an undesired artifact.  It does not show clustering structure within a group.
The second take is not bounded strictly but has a strong repulsion force to separate nodes of different neighbor groups.  The separation of hops is not obvious enough, especially for homogeneous graphs where the nodes are not shape coded.

The WebCola layout~\cite{dwyer2006ipsep} (Fig.~\ref{fig:design-alter}c) allows us to specify positional constraints between nodes, but due to its force-directed nature, it suffers from a similar problem as the D3 version in Fig.~\ref{fig:design-alter}a.

The space-filling curve layout~\cite{muelder2008rapid} is one of the fastest layout algorithms.  We choose a spiral curve as it can separate nodes of different groups from center to peripheral.  On a polar coordinate system, we use the topological distance as the distance on the curve between two consecutive nodes.  From Fig.~\ref{fig:design-alter}d, we can see that it has some spatial division between different groups but is even worse than the D3 version.  It also comes with an artifact that sometimes proximity in the polar coordinate system does not match the proximity in topology: nodes are placed with different radii but similar angles. Moreover, such a spiral space-filling design does not use space effectively, with a lot of unused space in the peripheral area where the less important neighbors reside.

\section{Results}\label{sec:results}

We evaluate CorGIE in two ways: we use it to address the questions in the two motivating scenarios, and we recruited \nt{five} GNN experts to participate in a user study to evaluate its utility and usability.

We first present our two usage scenarios. We then describe our study design, and discuss highlights of one of the expert user sessions using CorGIE on the popular Cora dataset.  We then summarize the study participants' feedback.  We describe \nt{another} expert user session in Supplemental Sec.~S4.4.

Below, we systematically describe four aspects of CorGIE usage: the \textit{visualization task} being conducted, the \textit{visual operation} used within the CorGIE interface, the \textit{direct observation} that is possible from the CorGIE display, and any resulting \textit{inference} of authors or participants. Here we tag only the \task{visualization task} with color; in Supplemental Sec.4 we provide a version of the text with all four aspects color-coded. 

\subsection{Usage scenario 1: Movie Recommendation}

\begin{figure}[!t]
  \centering
  \includegraphics[width=\linewidth]{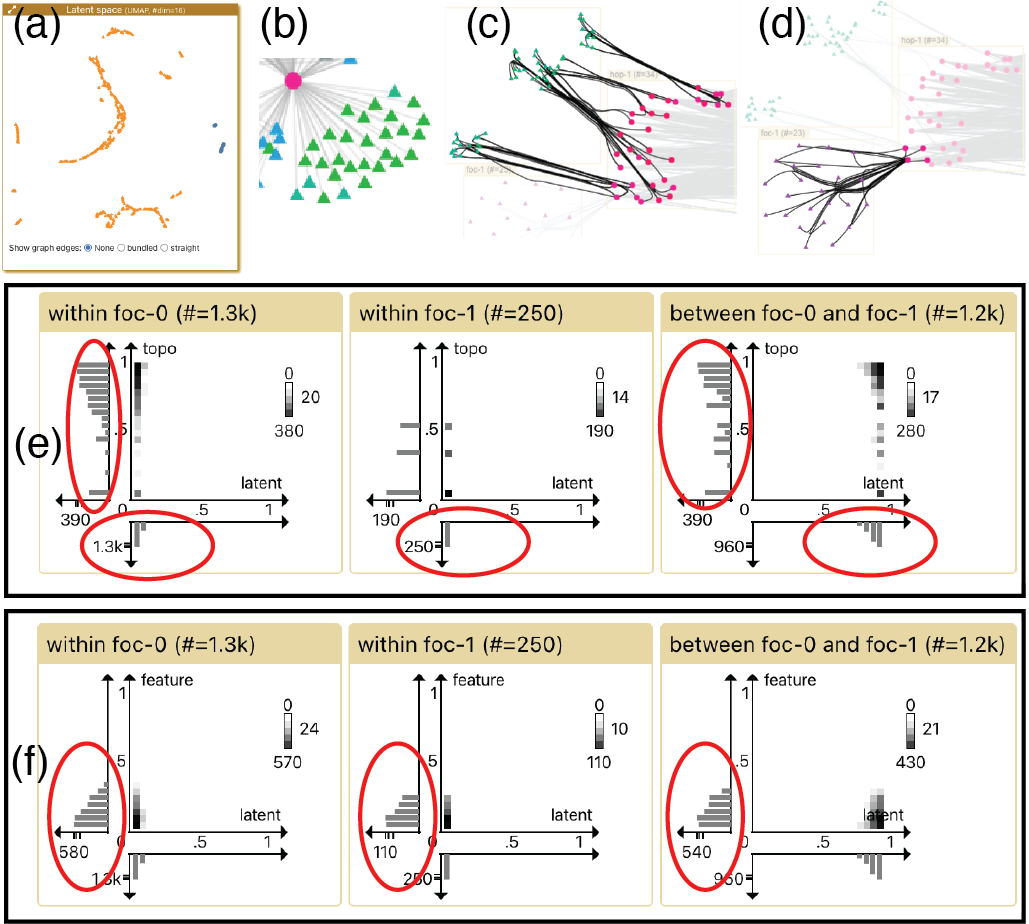}
  \caption[Check overview of the Movie scenario]{Check overview of the \textit{Movie} scenario: (a) \viewname{latent space}, movie nodes in blue and user nodes in orange; (b) inset of \viewname{global topology} showing similar peripheral nodes in similar green color (closeup of lower right for screenshot in \autoref{fig:teaser}).
    When comparing two clusters of user nodes, (c) we select nodes in \textit{foc-1} of the \viewname{K-hop topology}, highlighting themselves and four neighbors in the hop-1 box, and also in the (d) \viewname{global topology}.
    (e) We compare distances between topology and latent space, and (f) between feature and latent space.}
  \label{fig:movie-check-overview}
\end{figure}

To \task{obtain an overview} of the dataset, we first \op{color the nodes} by their node type (user and movie).  Fig.~\ref{fig:movie-check-overview}a shows that we can \ob{see} two clusters of movies in blue on the right-hand side and many clusters of users in orange in the rest of the latent space.
After \op{coloring the nodes} by the underlying positional colormap in the latent space, we \ob{observe}
in the \viewname{global topology} view (Fig.~\ref{fig:teaser}g) that most user nodes (triangles) are laid out around movie nodes (circles). Many nodes that are connected by only one edge are placed around the periphery of the layout, \infer{indicating} that there are many users who only watched one movie. In 
Fig.~\ref{fig:movie-check-overview}b, 
we \ob{notice} that the peripheral users that are connected to the same movies have similar colors: for example, the triangles on the bottom right are all green.
We can thus \infer{infer} that the GNN does a good job in grouping these one-time users reviewing the same movie.

To \task{compare user clusters}, we \op{select and focus on} two clusters from the \viewname{latent space}, one in the green zone (\textit{foc-0}) on the left, the other in the purple zone (\textit{foc-1}) on the upper right, as shown in Fig.~\ref{fig:teaser}.
In the \viewname{K-hop topology} (Fig.~\ref{fig:teaser}b), there \ob{appears} to be three green sub-clusters in \textit{foc-0}, and the green nodes connect to many nodes in the hop-1 box, while \textit{foc-1} does not show any salient internal structure.
To \task{understand the topological difference} between the two focal groups, we explore their connections in the \viewname{K-hop topology} with hover and select actions.  Fig.~\ref{fig:movie-check-overview}c and \ref{fig:movie-check-overview}d show the visual highlights in the \viewname{K-hop topology} when selecting the green \textit{foc-0} nodes and the purple \textit{foc-1} nodes respectively.
We can \ob{see} that the green nodes in the top \textit{foc-0} box connect to all hop-1 neighbors, but the purple nodes in the lower \textit{foc-1} box only connect to four of those neighbors.  We \infer{infer} that the GNN has learned the topological difference and thus separates them in the latent space. 

Besides graph topologies, we also want to \task{check whether distances match} between spaces.
In the \viewname{distance comparison view} (Fig.~\ref{fig:movie-check-overview}e), we can \ob{see} that the within group distances in latent space are small, while the between group distances are large, which \infer{confirms} that we have picked two distant clusters from the \viewname{latent space}.
We \ob{notice} that the topological distances within the green cluster (\textit{foc-0}) are relatively large considering they belong to the same cluster (leftmost vertical histogram in Fig.~\ref{fig:movie-check-overview}e), which actually \infer{matches} the existence of three sub-clusters in the \viewname{K-hop topology}. 
The feature distances in all three charts of Fig.~\ref{fig:movie-check-overview}f are \ob{small and similar}, \infer{indicating} that the node features (\#votes and average vote of a user) cannot distinguish \textit{foc-0} and \textit{foc-1}.  We further confirm the ineffectiveness of the features by \ob{reading} the feature distances by picking some other node groups (not shown in figure), which \infer{indicates} that they are not useful features, and could perhaps be removed from the dataset. 

\begin{figure*}[!t]
  \centering
  \includegraphics[width=\linewidth]{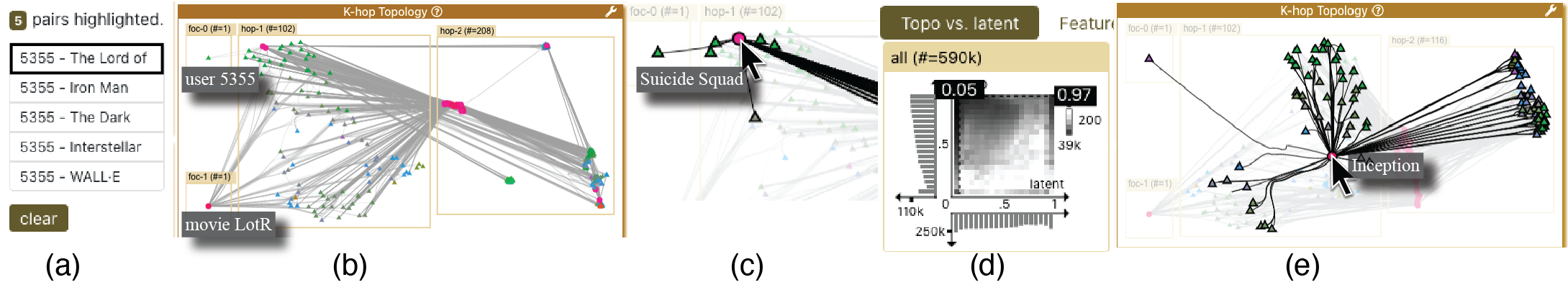}
  \caption[Check movie recommendations]{Check movie recommendations: (a) top 5 recommended movies for user \textit{5355}; (b) \viewname{K-hop topology} focusing user \textit{5355} and movie \textit{LotR}; (c) hover on the movie \textit{Suicide Squad} that the user has watched before; (d) topological distance between \textit{5355} and \textit{LotR} is large (\(0.97\)); (e) \viewname{K-hop topology} for user \textit{587} and movie \textit{LotR}.}
  \label{fig:movie-recomm}
\end{figure*}

After exploring the clustering structure, we \task{inspect instances} of recommendation. In Fig.~\ref{fig:movie-recomm}a we \op{select} the user node \textit{5355}, and \op{list} its top 5 recommended movies. The first one is \textit{The Lord of the Rings: the Return of King} (\textit{LotR}).
We would like to \task{understand why GNN decides to recommend \textit{LotR}}, so we \op{focus this recommendation} by clicking on it.  CorGIE automatically creates two focal groups with the user \textit{5355} in \textit{foc-0} and movie \textit{LotR} in \textit{foc-1} (Fig.~\ref{fig:movie-recomm}b).
The \viewname{K-hop topology} \ob{shows} that user \textit{5355} watched two movies (in pink) before.
When we \op{hover} on the two pink nodes in the hop-1 box (Fig.~\ref{fig:movie-recomm}c), we \ob{find out} that each only shares a few users with the recommended movie \textit{LotR}; for example, \textit{Suicide Squad} has only 5 hop-1 shared neighbors. 
We \infer{believe} that this recommendation is poor, potentially indicating that we \textit{are not there yet} with GNN training.   
We further confirmed this problem by \ob{reading} the topological \(0.05\) and latent \(0.97\) distances between user \textit{5355} and movie \textit{LotR} (Fig.~\ref{fig:movie-recomm}e), which \infer{seem} to be correlated negatively.

We later find a recommendation that makes sense: user \textit{587} and movie \textit{LotR}.  As shown in Fig.~\ref{fig:movie-recomm}e, the two movies that \textit{587} watched, \textit{Inception} and \textit{The Dark Night}, share many users with the recommended movie \textit{LotR} (many hop-1 neighbors are \ob{highlighted} when hovering on \textit{Inception}).  Moreover, the topological distance is \(0.72\) in this case, which is relatively small compared to the overall distance distribution (not shown in figure). 
CorGIE thus shows \infer{evidence} that this recommendation is well supported.

We repeat this process to \task{check other recommenda-} \task{tions}, and we find many that do not make sense.  We \infer{conclude} that this training result is not satisfactory.  It could be due to the ``cold start'' 
problem in such a small dataset, where most users only watch 1 or 2 movies.  Also, the node features were not very useful: e.g., we know that \#votes cannot distinguish users effectively.
To improve the recommendation, we might try a different model or fix the dataset problems.

\subsection{Usage scenario 2: Cora}

\begin{figure}[!t]
  \centering
  \includegraphics[width=\linewidth]{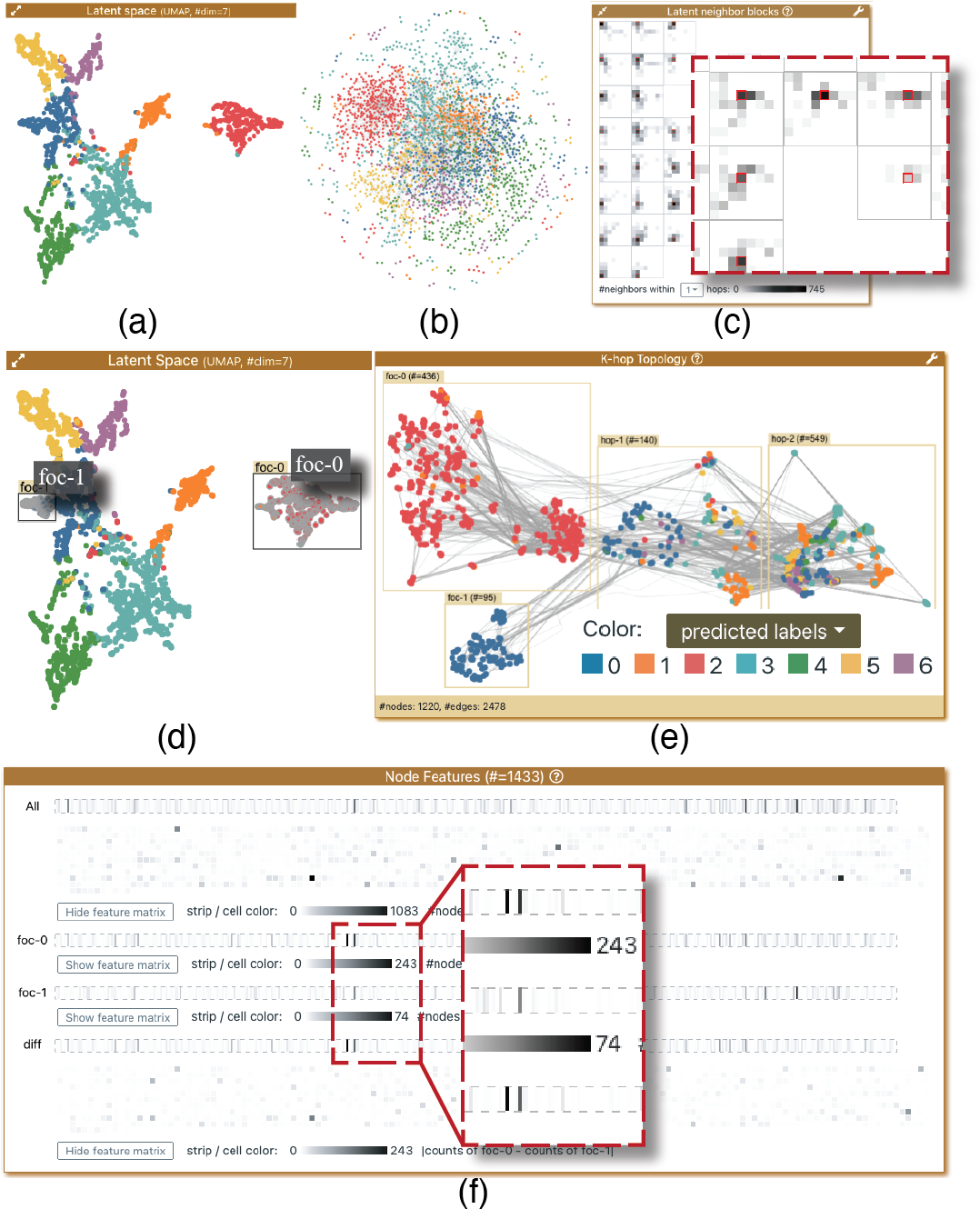}
  \caption[Explore overview of Cora dataset]{
    Explore overview of \textit{Cora} dataset, with nodes colored by predicted label: (a) \viewname{latent space}; (b) \viewname{global topology}; (c) \viewname{latent neighbor blocks} with zoomed-in inset showing that the red-outlined cells are darker than others.
    Compare two paper clusters: (d) focus on the red and blue clusters in the \viewname{latent space}; (e) \viewname{K-hop topology}; (f) \viewname{node features view} with inset showing the word count differences between \textit{foc-0} and \textit{foc-1}.}
  \label{fig:cora-overview}
\end{figure}

As with the \textit{Movie} scenario, we first \task{check the clustering} \task{structure} of paper nodes.
We \op{color the nodes by} the predicted labels to see if the label distribution makes sense.  In  \viewname{latent space} (Fig.~\ref{fig:cora-overview}a), we can \ob{see} that the different classes of papers are roughly separated to different areas.
In  \viewname{global topology} (Fig.~\ref{fig:cora-overview}b), although the force-directed layout lacks much structure, we can still \ob{see} that nearby nodes are in similar colors.
In \viewname{latent neighbor blocks} (Fig.~\ref{fig:cora-overview}c), we can \ob{see} that the red-outlined origin and its surrounding cells are darker for all the blocks.
All three observations \infer{indicate} that the GNN has done a good job in classifying papers using the graph topology.

Next, we \task{inspect and compare a few clusters}.  For example, when we compare \op{(select and focus)} the entire red cluster and a left part of the blue one (Fig.~\ref{fig:cora-overview}d), the \viewname{K-hop topology} (Fig.~\ref{fig:cora-overview}e) \ob{shows} that most nodes of the same color connect to each other, and the diff chart in the \viewname{node features view} (Fig.~\ref{fig:cora-overview}f) \ob{signals} a considerable amount of different words between the two paper clusters (many visible dark strips on the diff row).
These observations \infer{reinforce} our good impression on the training result. 

\begin{figure}[!t]
  \centering
  \includegraphics[width=\linewidth]{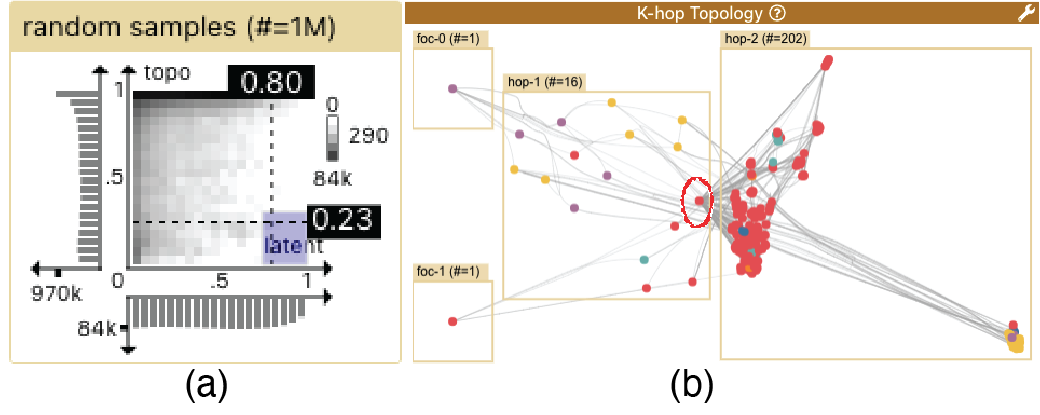}
  \caption[Find problematic node pairs in the Cora dataset]{Find problematic node pairs in the \textit{Cora} dataset: (a) select node pairs with high latent distance but low topological distance; (b) \viewname{K-hop topology} focusing on a buggy node pair, with only one shared hop-1 neighbor (circled) and many shared hop-2 neighbors.}
  \label{fig:cora-buggy-pair}
\end{figure}

As the GNN seems to classify most nodes successfully, we \task{look for problematic nodes and edges}.
We \op{brush} the distance scatterplot to highlight the node pairs with large latent distances but small topological distances (i.e. the bottom right area), as shown in Fig.~\ref{fig:cora-buggy-pair}a, and we \op{focus} one of these problematic node pairs.
In the \viewname{K-hop topology} (Fig.~\ref{fig:cora-buggy-pair}b), we \ob{find} through a few rounds of interactive \op{hover} that the two focal nodes only share one 1st-hop neighbor (circled in red), which has many nodes connected in the second hop.
As the Jaccard distance accounts for multiple hops of neighbors, the large number of shared hop-2 neighbors can explain the low topological distance (\(0.23\)).
We \infer{conjecture} that the GNN decides to locate them far from each other due to the large difference in the first hop.  Further investigation showed that the node in \textit{foc-0} is mis-classified, which \infer{hints} at the limitations of this GNN model to deal with such situations.

\subsection{Expert study overview}

\nt{All of the participants are actively working on GNN research.  They are an industrial lab researcher (P2), an undergraduate student (P4), and three senior PhD students (P1, P3, and P5).}
We had a meeting with \nt{P1 and P2} a few months before the study when we were iterating on the tasks and design of CorGIE.  In this meeting, we discussed the pain points in their workflow of model training and confirmed that our correspondence approach matches their mental model, but did not show any version of the CorGIE interface to the participants at that time.
In the remote study session, the first author introduced and demonstrated CorGIE for about 40 minutes.  Then participants used CorGIE on their own datasets \nt{or the demo ones} for about 30 minutes, followed by a semi-structured interview to gather their feedback. 
\nt{P1 and P2 used a near-final version of CorGIE, while the others used the final version.}

The results provide preliminary evidence of the utility of CorGIE in fulfilling the tasks we proposed in Sec.~\ref{sec:task-abs}, as the experts were highly positive about the effectiveness of CorGIE.

\subsection{Expert session: Cora decision boundary}

\begin{figure}[!t]
  \centering
  \includegraphics[width=\linewidth]{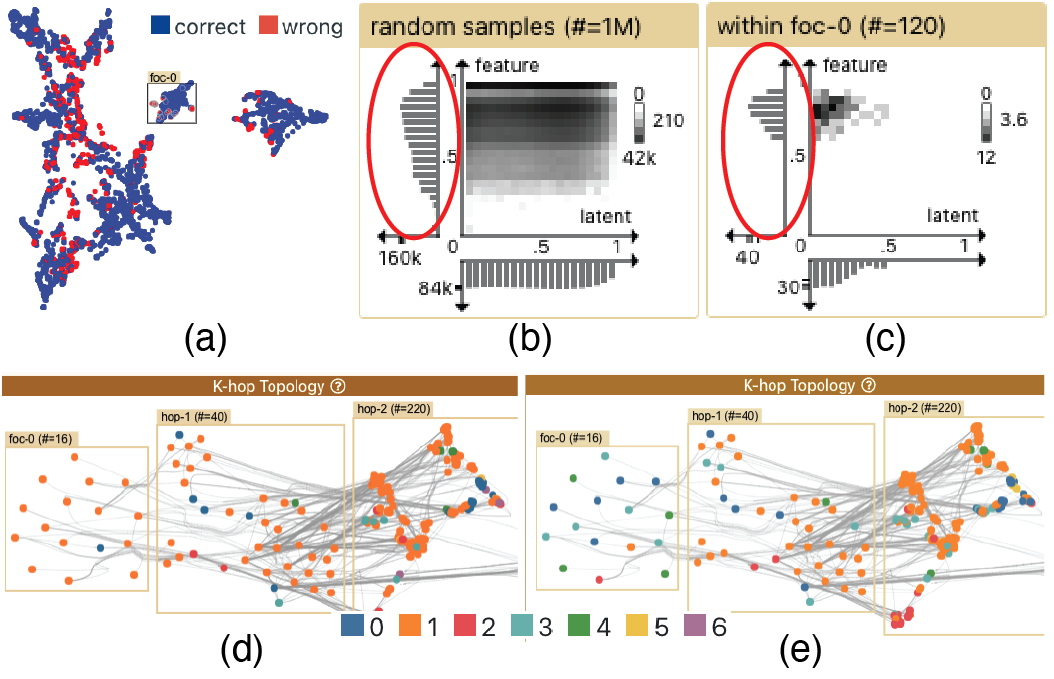}
  \caption[P1 explored the decision boundary in the Cora dataset]{P1 explored the decision boundary in the \textit{Cora} dataset: (a) he focused on the wrongly predicted nodes within a cluster (in the bounding box) in the \viewname{latent space}; he compared (b) feature distances of the entire graph to (c) those within the focused nodes; he inspected neighbors in the \viewname{K-hop topology} that are colored by (d) predicted label and (e) ground-truth label.}
  \label{fig:cora-case-study}
\end{figure}

In this case study, expert P1 first chose to perform quite similar tasks as in our usage scenario.  Then he wanted to \task{check the overview} to study misclassification.
He \op{colored the nodes by} label correctness (Fig.~\ref{fig:cora-case-study}a), where he \ob{saw} that the red (misclassified) nodes are distributed across different clusters.  He used a \op{filtered brush selection} to highlight and \op{focus} the wrongly predicted nodes within the middle cluster in the \viewname{latent space}.
He \op{colored the nodes by} the predicted labels (Fig.~\ref{fig:cora-case-study}d) and true labels (Fig.~\ref{fig:cora-case-study}e) respectively,
so he could \task{understand which classes are wrong}.  It \ob{appeared} that the GNN predicted these nodes as the orange class similar to their hop-1 neighbors, but the ground truth labels \ob{state} that they belong to multiple different classes (blue, green, red, etc) although the hop-1 neighbors are still orange.
Further inspection with \op{hover and brush selection} \op{shows} that the focal nodes are loosely connected to their hop-1 neighbors, but the hop-1 neighbors are tightly connected to the hop-2 ones.
Based on the variance in true labels and the sparsity in topology, P1 \infer{inferred} that these nodes were sitting at the decision boundary between two or multiple classes in the graph topology.  It was his first time to visually inspect the boundary nodes even though he had used this dataset for years.  He emphasized the importance of understanding the decision boundary if he needed to improve the GNN model.

After exploring the topology, P1 turned to  \viewname{distance comparison} for more information on the node features, where he \ob{found} the feature distances within the focal nodes are relatively large compared to the overall distribution (Fig.~\ref{fig:cora-case-study}c\&b).
He \infer{conjectured} that the current model performed as he expected, and that it would be unlikely to classify the boundary nodes correctly even if he kept training more epochs.
He also \infer{conjectured} that the GNN should use the node features in a better way to correct these nodes.
As he confirmed that the GNN has done a good job for most nodes except some boundary ones, he \infer{concluded} that there is no need to keep training more epochs, and training should be stopped to avoid over-fitting.

\subsection{Expert session feedback}
\nt{Overall, the participants were impressed after watching our demonstration and were able to use the tool themselves for the first time.
Here we summarize their feedback throughout the study.
Refer to the Supplemental Sec.~S5 for detailed interview questions and responses.}

\textbf{Tasks.} 
\nt{The participants} confirmed that the two-step (\textit{specify} and \textit{correspond}) task framework \nt{is useful in aiding the evaluation of GNNs, especially during the iterative refinement of focal nodes and when debugging node embeddings}. 
\nt{P1 noted that CorGIE can help find decision boundary nodes, a problem that people usually neglect and that was unsolvable problem for him before. He thought CorGIE could help him understand the data, the model, and the gap between the data and model. He thought he might be able to use the insight from CorGIE to improve his model. P2 found he could} debug models quickly in an intense leaderboard competition in ML research.

\textbf{Visual encoding.} 
\nt{The participants} stated that most views in CorGIE are straightforward to understand \nt{during} the introductory explanation.
\nt{Although all participants understood the intuition behind the
\viewname{node features} view, some thought they were unlikely to use it because they have no need to explore the distributions of an individual node feature.}
\nt{P3 noted that GNN researchers are not interested in the actual words in the Cora dataset as it is more of a natural language processing problem.
Two experts (P1 and P2) found the heat-map and heat-strips less intuitive without explanation.}

\textbf{K-hop layout.} 
\nt{All participants considered the \viewname{K-hop topology} to be intuitive, with the visualization aiding their reasoning of how a GNN functions. We consider this positive response a validation of the success of our novel K-hop layout algorithm.}

\textbf{Interaction.} 
\nt{All participants were able to pick up the main interactions quickly. They all verbally confirmed in the interview that the three-level interaction is easy to follow and facilitates the tasks well.}

\textbf{Usability.} 
\nt{Despite CorGIE being a research prototype, the participants managed to use it themselves successfully for most of the time during the remote studies. Nonetheless, certain usability issues came into our attention; 
for example, the mouse has to stay within the canvas area to perform a rectangular brush selection in the \viewname{latent space view}. The ability to indicate whether selection actions should affect only the target nodes themselves, or also their k-hop neighbors for a desired value of k, required explanation for four out of the five participants; this control was implemented using a drop-down menu in the \viewname{settings view}, which was easy to miss.} 

\nt{Participants P1 and P2, who used the near-final version of CorGIE, criticized the interface as being visually busy,} making it difficult to locate a specific object at first glance. 
\nt{We therefore improved the interface layouts to incorporate their feedback, for example by changing the important \textit{focus} action control into a very visually salient drop-down menu that looks like a corgi dog paw.}
\nt{P3, P4, and P5, who used the final version, were satisfied with the flow and visual complexity (P5: \textit{``This is much easier than writing Python scripts to analyze my training results''}).}

\nt{Overall, the expert feedback was positive that CorGIE is usable and learnable system. All participants expressed the excitement to incorporate CorGIE into their research workflow.}

\section{Related work}\label{sec:related-work}
CorGIE relates to previous research on explainable AI, which uses visualization to explain machine learning models.
In this section we relate and compare CorGIE to explainable visualizations for non-graph models, latent space, and graph neural networks.   We also discuss the related work for our new K-hop layout.

\subsection{Visualizations for machine learning models}
Many visualization tools have been developed to explain and evaluate machine learning models, especially in the recent five years.  Readers can refer to many survey papers to have a comprehensive understanding on this topic~\cite{chatzimparmpas2020star, hohman2019visual}.
Most such tools target models for images/videos and text sequences, most of which are not modeled as graphs. 
Although the seminal GNN work, GCN~\cite{kipf2016semi} stems from convolutional neural networks (CNNs) for images (as they both use convolution operation), it is not practical to simply retarget tools for CNNs to GNNs.  The key difference is that pixel neighbors in an image are intrinsically different from topological neighbors in a graph.  Sequence data like text in natural languages is even more different.
Unlike many visualizations that try to open the ``black box'' of a neural network and those that keep the ``black box'' completely opaque, we propose a ``grey box'' approach that only leverages a key concept in GNN - neighborhood aggregation (Sec.~\ref{sec:gnn-bg}) - to balance the generalizability and specificity 
for GNN models.

\subsection{Latent space interpreter}
The study of the latent space, also known as the embedding space, has attracted substantial attention, especially in text-based ML practices.  Embeddings eliminate the error-prone process of feature selection and they can be pre-trained for many different downstream applications~\cite{young2018recent}.
As CorGIE supports exploration of the correspondences between a latent space and a graph, it directly relates to much previous work on latent space visualization.

One theme is to reveal local and global structures of one or multiple latent spaces.
Ghosh et al.~\cite{ghosh2020visexpres} developed VisExPres, an interactive toolkit for user-driven evaluation of embeddings.  Heimerl et al.~\cite{heimerl2020embcomp} compare embeddings based on different quantitative metrics, while Cutura et al. does so with dimensional reduction techniques and matrix visualizations~\cite{cutura2020comparing}.  Liu et al.~\cite{liu2019latent} analogize the process of mapping and comparing semantic dimensions within latent spaces to latent space cartography.
This theme mainly focuses on digging deep inside latent spaces, which is different from our approach that tries to connect a latent space to the input.

Another theme, dimension reduction (DR), is also loosely related as many non-linear DR techniques produce latent spaces~\cite{nonato2018multidimensional}.
There are many visual tools for DR results: some for presenting the data points~\cite{smilkov2016embedding}, and some connect the DR results back to their semantically meaningful high-dimensional spaces~\cite{stahnke2016probing, faust2019dimreader}.
Their original input is usually tabular data, where a data item has multiple features (aka attributes), but none of them consider graph topology like CorGIE.

\subsection{GNN interpreter}
There has been some effort in the GNN community on GNN interpretation.  Most of the previous work focuses on generating algorithms or models to conduct feature and neighbor analysis~\cite{yuan2020explainability}.
Huang et al.~\cite{huang2020graphlime} propose GraphLIME to produce a few most representative features in the neighborhood of a node.
One of the most well-known work is the GNNExplainer~\cite{ying2019gnnexplainer} by Ying et al., who propose a method to compute the most important nodes of one or a group of user-specified nodes based on information theory.
Rao et al.~\cite{rao2020xfraud} follow up on GNNExplainer and propose xFraud to target at fraud detection specifically.
Pope et al.~\cite{pope2019explainability} extend explainability algorithms for CNN like saliency map, class activation mapping, and back-propagation to Graph-CNN.
This thread of work does not sufficiently support human-in-the-loop visual exploration, and so falls short in providing an overview of how well a GNN learns from the input graph and in connecting the exemplar inspections iteratively.

Two recent papers visualize graph models.
Li et al.~\cite{li2018embeddingvis} propose EmbeddingVis to compare multiple graph embeddings generated from different models.
It focuses on how the node metrics (e.g. degree, centrality) are preserved in the embeddings, but does not directly support topological neighborhood exploration nor node features.
As this research project was conducted before the recent trends of graph neural networks that use neighborhood aggregation, their embeddings are not generated by GNNs but other graph models.
A paper by Jin et al.~\cite{jin2020gnnvis}, who developed GNNVis to diagnose errors in GNN, is the most similar related work to CorGIE.
It only targets one downstream ML application, node classification, which limits its scope of usage.
Unlike CorGIE, GNNVis ignores the latent space and takes another route: it supports finding errors in the classification predictions, and comparing against two surrogate models to approximate the erroneous component in GNN.

\subsection{Graph layout}
We discuss the related work for our technical contribution, the K-hop layout.
A graph layout survey by Gibs et al.~\cite{gibson2013survey} categorizes graph layouts into three approaches: force-directed layouts, dimensional reduction layouts (e.g., MDS, t-SNE, UMAP), and multi-level graph layouts (see survey by McGee et al.~\cite{mcgee2019state}).
Our K-hop layout combines the DR and multi-level approaches.
Out of dozens of layout algorithms, ours is most related to those focusing on clusters.
IPSep-CoLa ~\cite{dwyer2006ipsep} by Dwyer et al. is a force-directed layout specializes at separation constraints.  Our investigation of its utility for this setting shows that it generates undesired artifacts (Fig.~\ref{fig:design-alter}c).
Our K-hop layout is similar to the Group-In-a-Box layout proposed by Rodrigues et al.~\cite{rodrigues2011group}
for category-based partitions of social networks.  It uses the space-filling treemap techniques to separate the clusters, whereas our division into groups is dynamic based on user selections and combines DR techniques.
The LinLog layout~\cite{noack2007energy} proposed by Noack is an energy-based model for cluster separation, and it inspires our readability metric in the adjustment step.

\section{Discussion and future work}

On reflection, after the whole process of characterizing the problem, designing the interface, and evaluating it, we believe that the high-level idea of exploring correspondences among input, output, and internal data structures 
is useful for GNN interpretation and may be generalizable to other deep neural networks. 
As a neural network itself is usually not interpretable for many reasons (e.g. non-linearlity, layering), the approach to ``open the black box'' completely demands substantial complexity of interpretation, and demands high expertise from the users.
We prefer a ``grey-box'' approach balancing the exposure of internal structure to a nonzero but minimum level.
Though often not exposed to the end users for interpretation, the node embedding is 
universal to all GNNs.
By finding correspondences between it and the input data (graph), we can infer if the GNN has achieved satisfactory results without fiddling around the internal structure of a neural network.
We imagine this approach could be generalized for other types of neural networks and ML applications.

We would also like to promote our notion of data spaces.  Our data abstraction consists of the three data spaces -- latent, topology, and feature -- 
alongside a task abstraction that connects them.  
This mental model of connecting data spaces helps us design the views: some are dedicated for a single space, while some connect multiple spaces.  
It also contributes to the usability and learnability of CorGIE, for which we have preliminary evidence through the feedback from \nt{five} GNN experts.  It would be interesting future work to introduce more data spaces, such as a geospatial space to deal with GNNs that specialize in geospatial data~\cite{guo2019attention}.

\nt{To evaluate CorGIE in realistic settings, we adhered to common practices in the GNN research community when training the datasets.  We made sure the hyperparameters were set properly and trained enough epochs until the models converge using quantitative metrics like loss value and classification accuracy. Our results show the value of CorGIE even when these practices appear to have succeeded: in the movie scenario where accuracy is over \(95\%\), we could still locate problematic nodes or \nt{recommendations} with CorGIE.}

\nt{CorGIE offers a visual way to explore local areas in a dataset, which is a time-consuming or even impossible process with other approaches. From our discussions with ML experts, we know that most GNN developers do not currently analyze their results in such a detailed way due to the lack of tools.
As shown in Sec.~\ref{sec:results}, CorGIE is particularly good at helping users generate hypotheses about the quality of the training results. Of course, verifying these hypotheses may require additional work outside the scope of this visual tool.}

It would be useful future work to handle larger graphs.  Due to the K-hop layout computation, the current version of CorGIE cannot guarantee a smooth interactive user experience for graphs with more than 20K nodes.
Although many popular benchmark datasets are within this scale, the ML community is moving forward with larger datasets, such as those on the OGB platform~\cite{hu2020open}.
\nt{It would be possible to accelerate the layout computation with a faster UMAP algorithm that leverages the GPU, which would be straighforward to add if a Javascript implementation becomes available.}
\nt{Another limitation for the K-hop layout is that a neighbor node that is both the first hop and the second hop of focal nodes could be only shown once in the first hop box; participant P5 noted this situation has the potential to mislead users.  
The Constellation system proposed to duplicate a node into a master and several proxies~\cite{munzner2000interactive} to address a similar issue, but that solution required significant interaction support to ensure that the node duplication was understandable. We chose not to do so because the CorGIE interface already has considerable visual complexity.} 

To further extend CorGIE,
we could incorporate the algorithms like GNNExplainer~\cite{ying2019gnnexplainer} either in the \viewname{K-hop topology view} implicitly or in a dedicated view explicitly.  This idea was also independently brought up by one of the GNN experts. We could also support more ways to specify nodes,  such as topological statistics. Finally, we could  enable the comparison of multiple node embeddings of the same input graph,
to help evaluate model re-architecting and hyper-parameter tuning.
A future paper could study how to enable GNN developers to explore the correspondences between multiple input graphs and their node/graph embeddings.

\section{Conclusion}
In this work, we present a task abstraction for exploring the correspondences between an input graph and the latent space created by a GNN, to understand if GNN has learned important characteristics from the graph and to find bugs in the latent space.
Based on this abstraction, we develop an interactive multi-view tool, CorGIE, which is validated through usage scenarios and case studies with GNN experts.
As the most important component in CorGIE, we propose the K-hop graph layout to reveal how GNNs aggregate information for nodes of interest. 
Both case studies and expert studies validate the effectiveness of bringing CorGIE into a GNN model development life-cycle. We envision that our novel data and task abstraction, in conjunction with our design rationales and implementation considerations, could serve as a stepping stone for future \nt{research}.

\ifCLASSOPTIONcompsoc
  \section*{Acknowledgments}
\else
  \section*{Acknowledgment}
\fi

The authors would like to thank Madison Elliott, Steve Kasica, Michael Oppermann, Ben Shneiderman, and Mara Solen for helpful comments on paper drafts, and the anonymous participants in the user study. 
\nt{We also thank Luyao Zhang for designing the logo of CorGIE.}
This work was funded in part by Uber Technologies and by NSERC RGPIN-2014-06309.  

\ifCLASSOPTIONcaptionsoff
  \newpage
\fi



\bibliographystyle{IEEEtran}
\bibliography{IEEEabrv, references}

\begin{thebibliography}{10}
\providecommand{\url}[1]{#1}
\csname url@samestyle\endcsname
\providecommand{\newblock}{\relax}
\providecommand{\bibinfo}[2]{#2}
\providecommand{\BIBentrySTDinterwordspacing}{\spaceskip=0pt\relax}
\providecommand{\BIBentryALTinterwordstretchfactor}{4}
\providecommand{\BIBentryALTinterwordspacing}{\spaceskip=\fontdimen2\font plus
\BIBentryALTinterwordstretchfactor\fontdimen3\font minus
  \fontdimen4\font\relax}
\providecommand{\BIBforeignlanguage}[2]{{%
\expandafter\ifx\csname l@#1\endcsname\relax
\typeout{** WARNING: IEEEtran.bst: No hyphenation pattern has been}%
\typeout{** loaded for the language `#1'. Using the pattern for}%
\typeout{** the default language instead.}%
\else
\language=\csname l@#1\endcsname
\fi
#2}}
\providecommand{\BIBdecl}{\relax}
\BIBdecl

\bibitem{dwivedi2020benchmarking}
V.~P. Dwivedi, C.~K. Joshi, T.~Laurent, Y.~Bengio, and X.~Bresson,
  ``Benchmarking graph neural networks,'' \emph{arXiv preprint
  arXiv:2003.00982}, 2020.

\bibitem{ying2019gnnexplainer}
R.~Ying, D.~Bourgeois, J.~You, M.~Zitnik, and J.~Leskovec, ``{GNNExplainer}:
  Generating explanations for graph neural networks,'' \emph{Advances in neural
  information processing systems (NeurIPS)}, vol.~32, pp. 9240--9251, 2019.

\bibitem{tang2015line}
J.~Tang, M.~Qu, M.~Wang, M.~Zhang, J.~Yan, and Q.~Mei, ``{LINE}: Large-scale
  information network embedding.'' in \emph{Proc. Intl. Conf. World Wide Web
  (WWW)}.\hskip 1em plus 0.5em minus 0.4em\relax ACM, 2015, p. 1067–1077.

\bibitem{Chami2020Machine}
I.~Chami, S.~Abu-El-Haija, B.~Perozzi, C.~R{\'e}, and K.~Murphy, ``Machine
  learning on graphs: A model and comprehensive taxonomy,'' \emph{ArXiv}, vol.
  abs/2005.03675, 2020.

\bibitem{peng2017cross}
N.~Peng, H.~Poon, C.~Quirk, K.~Toutanova, and W.-t. Yih, ``Cross-sentence n-ary
  relation extraction with graph {LSTM}s,'' \emph{Trans. Association for
  Computational Linguistics (ACL)}, vol.~5, pp. 101--115, 2017.

\bibitem{kipf2016semi}
T.~N. Kipf and M.~Welling, ``Semi-supervised classification with graph
  convolutional networks,'' \emph{arXiv preprint arXiv:1609.02907}, 2016.

\bibitem{ricci2011introduction}
F.~Ricci, L.~Rokach, and B.~Shapira, ``Introduction to recommender systems
  handbook,'' in \emph{Recommender systems handbook}.\hskip 1em plus 0.5em
  minus 0.4em\relax Springer, 2011, pp. 1--35.

\bibitem{mccallum2000automating}
A.~K. McCallum, K.~Nigam, J.~Rennie, and K.~Seymore, ``Automating the
  construction of internet portals with machine learning,'' \emph{Information
  Retrieval}, vol.~3, no.~2, pp. 127--163, 2000.

\bibitem{velivckovic2017graph}
P.~Veli{\v{c}}kovi{\'c}, G.~Cucurull, A.~Casanova, A.~Romero, P.~Lio, and
  Y.~Bengio, ``Graph attention networks,'' \emph{arXiv preprint
  arXiv:1710.10903}, 2017.

\bibitem{yang2020heterogeneous}
C.~Yang, Y.~Xiao, Y.~Zhang, Y.~Sun, and J.~Han, ``Heterogeneous network
  representation learning: A unified framework with survey and benchmark,''
  \emph{IEEE Trans. Knowledge and Data Engineering (TKDE)}, 2020.

\bibitem{mcinnes2018umap}
L.~McInnes, J.~Healy, and J.~Melville, ``{UMAP}: Uniform manifold approximation
  and projection for dimension reduction,'' \emph{arXiv preprint
  arXiv:1802.03426}, 2018.

\bibitem{bernard2015survey}
J.~Bernard, M.~Steiger, S.~Mittelst{\"a}dt, S.~Thum, D.~Keim, and
  J.~Kohlhammer, ``A survey and task-based quality assessment of static {2D}
  colormaps,'' in \emph{Visualization and Data Analysis 2015}, vol. 9397.\hskip
  1em plus 0.5em minus 0.4em\relax International Society for Optics and
  Photonics, 2015, p. 93970M.

\bibitem{bostock2011d3}
M.~Bostock, V.~Ogievetsky, and J.~Heer, ``{D$^3$} data-driven documents,''
  \emph{IEEE Trans. Visualization and Computer Graphics (TVCG)}, vol.~17,
  no.~12, pp. 2301--2309, 2011.

\bibitem{wood2010visualisation}
J.~Wood, J.~Dykes, and A.~Slingsby, ``Visualisation of origins, destinations
  and flows with {OD} maps,'' \emph{The Cartographic Journal}, vol.~47, no.~2,
  pp. 117--129, 2010.

\bibitem{lex2014upset}
A.~Lex, N.~Gehlenborg, H.~Strobelt, R.~Vuillemot, and H.~Pfister, ``Up{S}et:
  Visualization of intersecting sets,'' \emph{IEEE Trans. Visualization and
  Computer Graphics (TVCG)}, vol.~20, no.~12, pp. 1983--1992, 2014.

\bibitem{van2008visualizing}
L.~Van~der Maaten and G.~Hinton, ``Visualizing data using {t-SNE},''
  \emph{Journal of machine learning research}, vol.~9, no.~11, 2008.

\bibitem{coenen2019understanding}
A.~Coenen and A.~Pearce, ``Understanding {UMAP},''
  \url{https://pair-code.github.io/understanding-umap/}, 2019.

\bibitem{borg2005modern}
I.~Borg and P.~J. Groenen, \emph{Modern multidimensional scaling: Theory and
  applications}.\hskip 1em plus 0.5em minus 0.4em\relax Springer Science \&
  Business Media, 2005.

\bibitem{noack2007energy}
A.~Noack, ``Energy models for graph clustering.'' \emph{Journal of Graph
  Algorithms and Applications}, vol.~11, no.~2, pp. 453--480, 2007.

\bibitem{zhou2013edge}
H.~Zhou, P.~Xu, X.~Yuan, and H.~Qu, ``Edge bundling in information
  visualization,'' \emph{Tsinghua Science and Technology}, vol.~18, no.~2, pp.
  145--156, 2013.

\bibitem{gansner2011multilevel}
E.~R. Gansner, Y.~Hu, S.~North, and C.~Scheidegger, ``Multilevel agglomerative
  edge bundling for visualizing large graphs,'' in \emph{2011 IEEE Pacific
  Visualization Symposium}, pp. 187--194.

\bibitem{tang2008arnetminer}
J.~Tang, J.~Zhang, L.~Yao, J.~Li, L.~Zhang, and Z.~Su, ``Arnetminer: Extraction
  and mining of academic social networks,'' in \emph{Proc. 14th ACM SIGKDD
  Intl. Conf. Knowledge Discovery and Data Mining (KDD)}, New York, NY, USA,
  2008, p. 990–998.

\bibitem{dwyer2006ipsep}
T.~Dwyer, Y.~Koren, and K.~Marriott, ``{IPSep-CoLa}: An incremental procedure
  for separation constraint layout of graphs,'' \emph{IEEE Trans. Visualization
  and Computer Graphics (TVCG)}, vol.~12, no.~5, pp. 821--828, 2006.

\bibitem{muelder2008rapid}
C.~Muelder and K.-L. Ma, ``Rapid graph layout using space filling curves,''
  \emph{IEEE Trans. Visualization and Computer Graphics (TVCG)}, vol.~14,
  no.~6, pp. 1301--1308, 2008.

\bibitem{chatzimparmpas2020star}
A.~Chatzimparmpas, R.~M. Martins, I.~Jusufi, K.~Kucher, F.~Rossi, and
  A.~Kerren, ``The state of the art in enhancing trust in machine learning
  models with the use of visualizations,'' \emph{Computer Graphics Forum},
  vol.~39, no.~3, pp. 713--756, 2020.

\bibitem{hohman2019visual}
F.~Hohman, M.~Kahng, R.~Pienta, and D.~H. Chau, ``Visual analytics in deep
  learning: An interrogative survey for the next frontiers,'' \emph{IEEE Trans.
  Visualization and Computer Graphics (TVCG)}, vol.~25, no.~8, pp. 2674--2693,
  2019.

\bibitem{young2018recent}
T.~Young, D.~Hazarika, S.~Poria, and E.~Cambria, ``Recent trends in deep
  learning based natural language processing,'' \emph{IEEE Computational
  intelligence magazine}, vol.~13, no.~3, pp. 55--75, 2018.

\bibitem{ghosh2020visexpres}
A.~Ghosh, M.~Nashaat, J.~Miller, and S.~Quader, ``Vis{E}x{P}re{S}: A visual
  interactive toolkit for user-driven evaluations of embeddings,'' \emph{IEEE
  Trans. Visualization and Computer Graphics (TVCG)}, pp. 1--1, 2020.

\bibitem{heimerl2020embcomp}
F.~Heimerl, C.~Kralj, T.~Moller, and M.~Gleicher, ``embcomp: Visual interactive
  comparison of vector embeddings,'' \emph{IEEE Trans. Visualization and
  Computer Graphics (TVCG)}, 2020.

\bibitem{cutura2020comparing}
R.~Cutura, M.~Aupetit, J.-D. Fekete, and M.~Sedlmair, ``Comparing and exploring
  high-dimensional data with dimensionality reduction algorithms and matrix
  visualizations,'' in \emph{Proc. Intl. Conf. Advanced Visual Interfaces
  (AVI)}.\hskip 1em plus 0.5em minus 0.4em\relax ACM, 2020.

\bibitem{liu2019latent}
Y.~Liu, E.~Jun, Q.~Li, and J.~Heer, ``Latent space cartography: Visual analysis
  of vector space embeddings,'' \emph{Computer Graphics Forum}, vol.~38, pp.
  67--78, 06 2019.

\bibitem{nonato2018multidimensional}
L.~G. Nonato and M.~Aupetit, ``Multidimensional projection for visual
  analytics: Linking techniques with distortions, tasks, and layout
  enrichment,'' \emph{IEEE Trans. Visualization and Computer Graphics (TVCG)},
  vol.~25, no.~8, pp. 2650--2673, 2018.

\bibitem{smilkov2016embedding}
D.~Smilkov, N.~Thorat, C.~Nicholson, E.~Reif, F.~B. Vi{\'e}gas, and
  M.~Wattenberg, ``Embedding projector: Interactive visualization and
  interpretation of embeddings,'' \emph{arXiv preprint arXiv:1611.05469}, 2016.

\bibitem{stahnke2016probing}
J.~Stahnke, M.~Dörk, B.~Müller, and A.~Thom, ``Probing projections:
  Interaction techniques for interpreting arrangements and errors of
  dimensionality reductions,'' \emph{IEEE Trans. Visualization and Computer
  Graphics (TVCG)}, vol.~22, no.~1, pp. 629--638, 2016.

\bibitem{faust2019dimreader}
R.~Faust, D.~Glickenstein, and C.~Scheidegger, ``Dim{R}eader: Axis lines that
  explain non-linear projections,'' \emph{IEEE Trans. Visualization and
  Computer Graphics (TVCG)}, vol.~25, no.~1, pp. 481--490, 2019.

\bibitem{yuan2020explainability}
H.~Yuan, H.~Yu, S.~Gui, and S.~Ji, ``Explainability in graph neural networks: A
  taxonomic survey,'' \emph{arXiv preprint arXiv:2012.15445}, 2020.

\bibitem{huang2020graphlime}
Q.~Huang, M.~Yamada, Y.~Tian, D.~Singh, D.~Yin, and Y.~Chang, ``Graph{LIME}:
  Local interpretable model explanations for graph neural networks,''
  \emph{arXiv preprint arXiv:2001.06216}, 2020.

\bibitem{rao2020xfraud}
S.~X. Rao, S.~Zhang, Z.~Han, Z.~Zhang, W.~Min, Z.~Chen, Y.~Shan, Y.~Zhao, and
  C.~Zhang, ``{xFraud}: Explainable fraud transaction detection on
  heterogeneous graphs,'' \emph{arXiv preprint arXiv:2011.12193}, 2020.

\bibitem{pope2019explainability}
P.~E. Pope, S.~Kolouri, M.~Rostami, C.~E. Martin, and H.~Hoffmann,
  ``Explainability methods for graph convolutional neural networks,'' in
  \emph{IEEE/CVF Conf. Computer Vision and Pattern Recognition (CVPR)}, 2019,
  pp. 10\,764--10\,773.

\bibitem{li2018embeddingvis}
Q.~Li, K.~S. Njotoprawiro, H.~Haleem, Q.~Chen, C.~Yi, and X.~Ma,
  ``Embedding{V}is: A visual analytics approach to comparative network
  embedding inspection,'' in \emph{IEEE Conf. Visual Analytics Science and
  Technology (VAST)}, 2018, pp. 48--59.

\bibitem{jin2020gnnvis}
Z.~Jin, Y.~Wang, Q.~Wang, Y.~Ming, T.~Ma, and H.~Qu, ``{GNNVis}: A visual
  analytics approach for prediction error diagnosis of graph neural networks,''
  \emph{arXiv preprint arXiv:2011.11048}, 2020.

\bibitem{gibson2013survey}
H.~Gibson, J.~Faith, and P.~Vickers, ``A survey of two-dimensional graph layout
  techniques for information visualisation,'' \emph{Information Visualization},
  vol.~12, no. 3-4, pp. 324--357, 2013.

\bibitem{mcgee2019state}
F.~McGee, M.~Ghoniem, G.~Melan{\c{c}}on, B.~Otjacques, and B.~Pinaud, ``The
  state of the art in multilayer network visualization,'' in \emph{Computer
  Graphics Forum}, vol.~38, no.~6.\hskip 1em plus 0.5em minus 0.4em\relax Wiley
  Online Library, 2019, pp. 125--149.

\bibitem{rodrigues2011group}
E.~M. Rodrigues, N.~Milic-Frayling, M.~Smith, B.~Shneiderman, and D.~Hansen,
  ``Group-in-a-box layout for multi-faceted analysis of communities,'' in
  \emph{IEEE Intl. Conf. Privacy, Security, Risk and Trust (PASSAT) and IEEE
  Intl. Conf. Social Computing (SocialCom)}, 2011, pp. 354--361.

\bibitem{guo2019attention}
S.~Guo, Y.~Lin, N.~Feng, C.~Song, and H.~Wan, ``Attention based
  spatial-temporal graph convolutional networks for traffic flow forecasting,''
  in \emph{Proc. AAAI Conference on Artificial Intelligence}, vol.~33, no.~01,
  2019, pp. 922--929.

\bibitem{hu2020open}
W.~Hu, M.~Fey, M.~Zitnik, Y.~Dong, H.~Ren, B.~Liu, M.~Catasta, and J.~Leskovec,
  ``Open graph benchmark: Datasets for machine learning on graphs,''
  \emph{arXiv preprint arXiv:2005.00687}, 2020.

\bibitem{munzner2000interactive}
T.~M. Munzner, \emph{Interactive visualization of large graphs and
  networks}.\hskip 1em plus 0.5em minus 0.4em\relax Stanford University, 2000.

\end{thebibliography}
%

%







\begin{IEEEbiography}[{\includegraphics[width=1in,height=1.25in,clip,keepaspectratio]{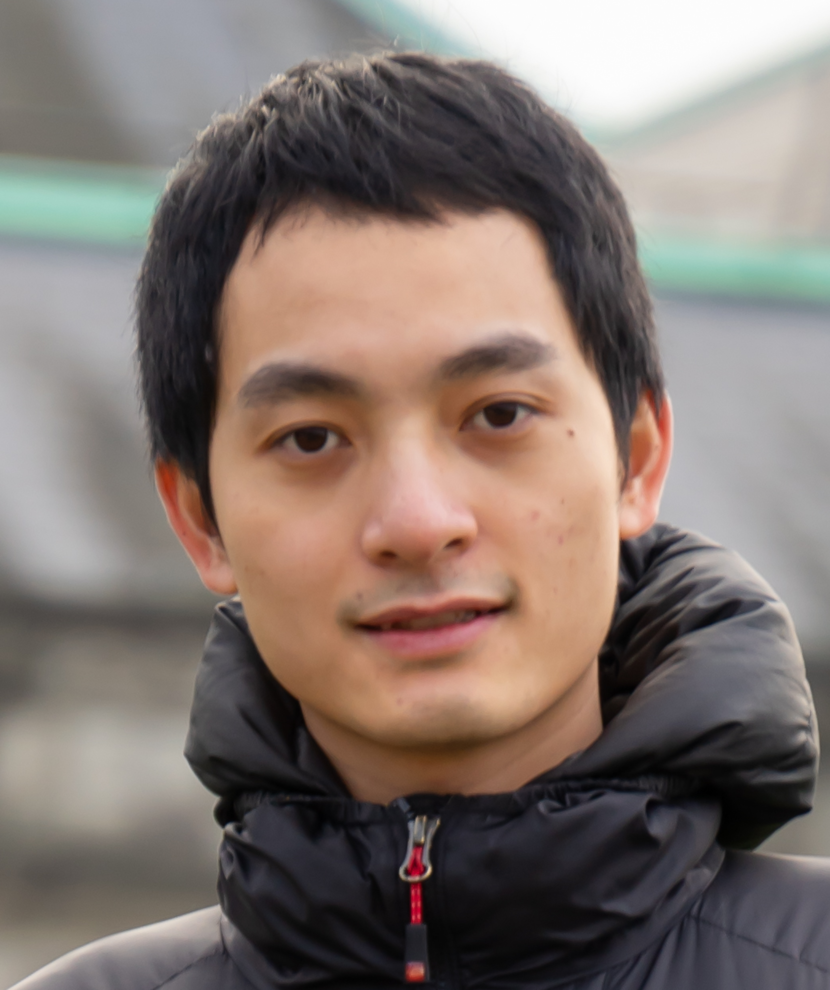}}]{Zipeng Liu}
received the PhD degree from University of British Columbia in 2021, and the BS degree from Peking University in 2015. His research involves visualization of multi-level structures in trees, graphs, logs and machine learning models.
\end{IEEEbiography}

\begin{IEEEbiography}[{\includegraphics[width=1in,height=1.25in,clip,keepaspectratio]{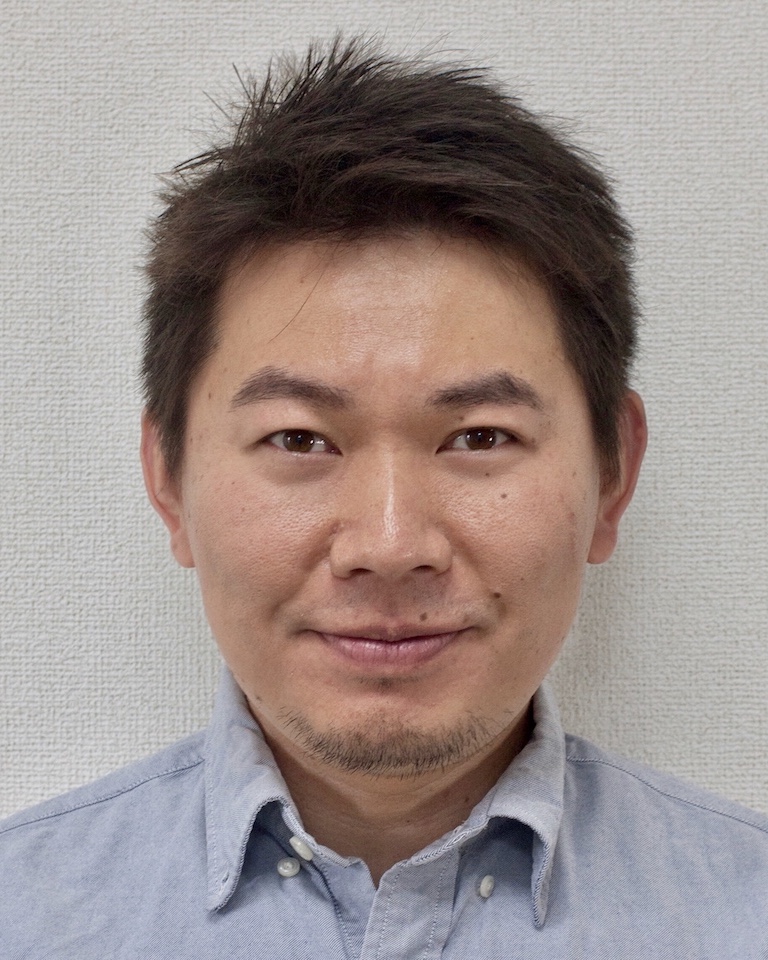}}]{Yang Wang}
is a Research Scientist Manager at Facebook supporting the AI-aided Creative Optimization domain. His research focuses on Visual Analytics, Human-Computer Interaction, and Machine Learning. Specifically, methodologies and systems to help users better understand and leverage machine learning in their analytical and creative processes.

\end{IEEEbiography}

\begin{IEEEbiography}[{\includegraphics[width=1in,height=1.25in,clip,keepaspectratio]{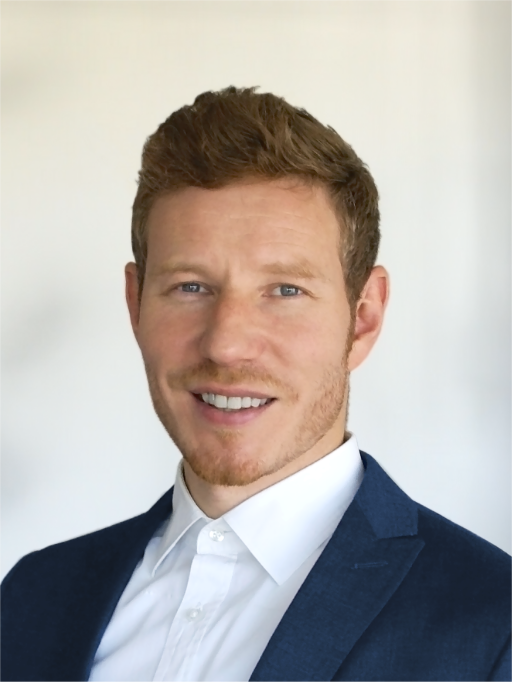}}]{J{\"u}rgen Bernard}
is an Assistant Professor at the University of Zurich and head of the Interactive Visual Data Analysis group. He has received the Ph.D degree in computer science from the University of Darmstadt, and was a postdoctoral research fellow at the University of British Columbia, Canada. His research focus is on visual analytics and human-centered  machine learning.
\end{IEEEbiography}

\begin{IEEEbiography}[{\includegraphics[width=1in,height=1.25in,clip,keepaspectratio]{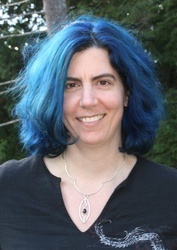}}]{Tamara Munzner}
(Senior Member, IEEE) received the PhD degree from Stanford. She is currently a professor with the University of British Columbia. She worked on visualization projects in a broad range of application domains from genomics to journalism. Her book Visualization Analysis and Design appeared in 2014. She was the recipient of the IEEE VGTC Visualization Technical Achievement Award.
\end{IEEEbiography}

\end{document}